\pdfoutput=1
\documentclass[11pt]{article}

\usepackage[final]{acl}

\usepackage{times}
\usepackage{latexsym}

\usepackage[T1]{fontenc}
\usepackage{tabularx}
\usepackage{longtable}
\usepackage{booktabs}
\usepackage{graphicx}
\usepackage{rotating}
\usepackage{multirow}
\usepackage{comment}

\usepackage[utf8]{inputenc}

\usepackage{microtype}

\usepackage{inconsolata}

\usepackage{graphicx}
\usepackage{xcolor}

\definecolor{massive_benchmarks}{RGB}{99,172,91}
\definecolor{interactive_games}{RGB}{114,75,164}

\newcommand{\cut}[1]{}
\newcounter{tbsnr}
\newenvironment{tbs}
{\addtocounter{tbsnr}{1}\par\bigskip\noindent\fbox{\thetbsnr}
\hspace*{\fill}\begin{minipage}{7cm}\tt}
{\end{minipage}\hspace*{\fill}\bigskip}

\usepackage[color=lightgray]{todonotes}
\usepackage{enumitem}
\usepackage{booktabs}

\definecolor{Green}{RGB}{98,115,19} 
\usepackage{soul}
\sethlcolor{lightgray}

\usepackage{amsmath}

\title{Triangulating LLM Progress\\ through Benchmarks, Games, and Cognitive Tests}

\author{
  \textbf{Filippo Momentè\textsuperscript{1}},
  \textbf{Alessandro Suglia\textsuperscript{2}\thanks{Work done while at Heriot-Watt University.}},
  \textbf{Mario Giulianelli\textsuperscript{3}},
  \textbf{Ambra Ferrari\textsuperscript{1}},
\\
  \textbf{Alexander Koller\textsuperscript{4}},
  \textbf{Oliver Lemon\textsuperscript{5}},
  \textbf{David Schlangen\textsuperscript{6,7}}\\
  \textbf{Raquel Fernández\textsuperscript{8}},
  \textbf{Raffaella Bernardi\textsuperscript{9}\thanks{Work partially done while at University of Trento.}} \vspace{0.5em}
\\
\textsuperscript{1}University of Trento,
  \textsuperscript{2}ILCC University of Edinburgh,
  \textsuperscript{3}UCL,\\
  \textsuperscript{4}Saarland University,
  \textsuperscript{5}Heriot-Watt University,
  \textsuperscript{6}University of Potsdam,
  \textsuperscript{7}DFKI,\\
  \textsuperscript{8}University of Amsterdam,
    \textsuperscript{9}Faculty of Engineering, Free University of Bozen-Bolzano\\
\texttt{\href{mailto:lm-playschool@googlegroups.com}{lm-playschool@googlegroups.com}}
}

\begin{document}
\maketitle
%--------------------------------------------
\begin{abstract}
% We examine three evaluation paradigms for LLMs and argue that they provide complementary perspectives. 
\begin{comment} % previous abstract (14 Feb 9pm CET)
We investigate which of two evaluation paradigms---large static question-answering benchmarks (e.g., MMLU and BBH) or interactive games (e.g., Signalling Games or Taboo)---is most effective at discriminating %differentiating %performance across 
LLMs of varying quality.
Then, inspired by human cognitive assessments, we compile a suite of targeted tests %focus on evaluation datasets 
that measure cognitive abilities deemed essential for effective language use and investigate their correlation with model performance in benchmarks and games. 
Our analyses reveal that interactive games are superior to static benchmarks in discriminating models.
Peformance in games correlates with common-sense reasoning, working memory, and theory of mind, while performance on static benchmarks correlates more strongly with reasoning, planning, and pragmatic skills. %\todo{\rb{This sentence about correlation is misleading. MML and BBH have higher correlation with causal/logical reasoning and even planning than the games}}.
We advocate the development of new interactive benchmarks and targeted cognitive tasks inspired by assessing human abilities but designed specifically for LLMs.\looseness-1  
% We argue for future work in developing interactive benchmarks and carefully designed tasks inspired by assessing human cognitive abilities.  
\end{comment}
% \begin{comment}  % New attempt
We examine three evaluation paradigms: standard 
%general reference-based 
benchmarks (e.g., MMLU and BBH), interactive games (e.g., Signalling Games or Taboo), and cognitive tests (e.g., for working memory or theory of mind).
First, we investigate which of the former two---benchmarks or games---is most effective at discriminating LLMs of varying quality.
Then, inspired by human cognitive assessments, we compile a suite of targeted tests that measure cognitive abilities deemed essential for effective language use, and we investigate their correlation with model performance in benchmarks and games. 
Our analyses reveal that interactive games are superior to standard benchmarks in discriminating models.
Causal and logical reasoning correlate with both static and interactive tests, while differences emerge regarding core executive functions and social/emotional skills, which correlate more with games.
%in particular working memory and emotional intelligence are only significantly correlated with performance in games.
%Performance in games correlates with common-sense reasoning, working memory, and theory of mind, while performance on static benchmarks correlates more strongly with reasoning, planning, and pragmatic skills.
We propose a new evaluation framework triangulating LLMs progress. Our findings highlight the importance of defining evaluation regimes that consider multiple paradigms. We advocate for the development of new interactive benchmarks and targeted cognitive tasks inspired by human ability assessments but designed specifically for LLMs.
The code for running the experiments is released at: \url{https://github.com/momentino/playpen_eval/tree/triangulating}.

% \end{comment}
\end{abstract}
%--------------------------------------------

%====================================
\section{Introduction}
\label{sec:intro}
%====================================

Evaluating LLMs is critical to track progress, identify blind spots, and ultimately advance towards the kind of language-based AI systems we want as a society~\cite{wooldridge1995intelligent}. Currently, 
the most widespread way to evaluate LLMs is by means of \textbf{large benchmarks} made up of miscellaneous 
%question-answering (QA) 
tasks. 
Pre-LLM benchmarks such as GLUE and SuperGLUE~\cite{glue,NEURIPS2019_4496bf24} have been replaced by even larger evaluation suites such as MMLU~\cite[Measuring Massive Multitask Language Understanding;][]{hendryckstest2021}, MMLU-Pro ~\cite{mmlupro}, or BBH~\cite[BIG-Bench Hard;][]{suzgun-etal-2023-challenging,srivastava2023beyond-short}. Models with high performance on these benchmarks are taken to possess extensive \textbf{world knowledge along with complex problem-solving abilities.} 

This trend has promoted standardisation in LLM evaluation protocols, with online leaderboards constantly updated as new models are released. 
Despite this undeniable benefit, general-purpose  
%reference-based 
 benchmarks, like those mentioned above, are not without problems. 
Evaluation results may be inflated by data contamination (see, e.g.,~\citealt{gema:arewe25} for MMLU) and distorted by model sensitivity to prompt format~\cite{zhuo2024prosa}. 
Moreover, by design, such benchmarks
overlook actual language use in favour of knowledge-intensive tasks where success is measured against gold %-standard 
reference answers provided in a single conversational turn. 

This contrasts with the view, put forward by philosophers and psycholinguists alike \cite{wittgenstein1953investigations,austin1962things,searle1969speech,clark1996using}, that the quintessence of language resides in \textit{situated language use}, i.e., using language for a purpose in social and task-based multi-turn interactions \cite{bisk-etal-2020-experience}. %\looseness-1
This situated and interactive view underpins a parallel %evaluation 
trend where LLMs are evaluated as \textbf{goal-directed language users} by means of \textbf{interactive games}~\cite{bertolazzi-etal-2023-chatgpts,schla:what23,suglia2024visually}.\footnote{Online leaderboards have started to appear for the interactive games evaluation paradigm; see, e.g., \url{https://textarena.ai/}, \url{https://clembench.github.io}.} 
This interactive evaluation paradigm goes beyond single-turn text generation, which is critical for deploying LLMs as agents. %, and is less prone to data contamination.  
Additionally, it is less susceptible to data contamination
%---both because LLM training data primarily comes from web sources and 
because the vast space of possible multi-turn interactions is unlikely to be fully represented in the training data. 
%As a result, interactive games provide a more robust framework for evaluating the true generalisation capacity of LLMs~\cite{hupkes2023taxonomy}.
Yet, despite these advantages, it is not easy to pinpoint which specific abilities underpin models' performance on interactive language games---a difficulty that to some extent also applies to static question-answering benchmarks such as MMLU.\looseness-1

In this paper, we study these two evaluation paradigms---general-purpose %reference-based 
benchmarks and interactive games---and show that they can provide complementary perspectives. 
%First, 
In particular,
we investigate whether %the  comprehensive static 
standard benchmarks or games are more effective in gauging qualitative differences between models, %, e.g., % between models, 
%across model families and sizes, 
evaluate a selection of current LLMs from five %different 
model families, and
%We 
find that %interactive 
games highlight more nuanced differences between %current
LLMs compared to %static %reference-based 
standard benchmarks. While scaling model size leads to systematic improvements on %static 
benchmarks, game-based evaluation reveals differences between models of the same size.
%it doesn't guarantee performance boosts in interactive language use. 

To shed light on the abilities underlying models' performance on these two evaluation frameworks, we resort to \textbf{targeted cognitive tests}. In the absence of specifically designed tests for LLMs, we %We 
propose a taxonomy of cognitive skills motivated by human neurocognitive science and compile a list of existing evaluation datasets designed to assess each skill in isolation. 
Specifically, we investigate to what extent increased performance on specific abilities correlates with performance gain in standard %the general-purpose reference-based 
benchmarks and interactive games, respectively.\looseness-1  

In sum, our work brings in a unique perspective by studying the interrelations between a diverse set of evaluation paradigms which, to our knowledge, have not been jointly investigated before. 

% Our analysis shows that while %higher-order 
% causal and logical reasoning %abilities
% correlate with both static and interactive tests, differences emerge regarding core executive functions and social/emotional skills; in particular, working memory and emotional intelligence %abilities 
% are only significantly correlated with performance in%interactive 
% games.

%====================================
\section{Models}
\label{sec:models}
%====================================
 
We apply our evaluation framework to the following open-weight LLMs ranging from 3.8B to 72B models:  
Phi-4-Mini with 3.8B parameters \cite{phi4}, 
OLMo-2-1124 with 7B and 13B parameters \cite{olmo20242olmo2furious},  
Falcon3 with 10B parameters \cite{falcon3}, 
Qwen2.5 with 7B, 32B, and 72B parameters \cite{qwen2, qwen2.5}, and 
Llama-3 with 8B and 70B parameters \cite{dubey2024llama3herdmodels}. 
All these models score above 70\% on IFEval~\cite{zhou2023instruction}, %which indicates t
which we take as indication that they possess the instruction-following capabilities necessary for playing games and understanding task descriptions in the cognitive tests. See Appendix~\ref{sec:appendix-models} for further model details.

%====================================
\section{How to Identify Blind Spots in LLMs}
\label{sec:static-interactive}
%====================================

Evaluation paradigms have most practical use when they allow us to track progress by identifying blind spots in LLMs. Here we compare the two paradigms under study on the extent to which they highlight differences between current models, helping us form hypotheses about possible performance gaps and successful mitigation strategies.

\paragraph{General-purpose reference-based benchmarks}  We take MMLU \cite{hendryckstest2021} and BBH~\cite{suzgun-etal-2023-challenging} as representative of standard
%general-purpose reference-based 
benchmarks. MMLU evaluates whether LLMs can apply knowledge from specific domains: it consists of multiple-choice questions spanning 57 academic subjects. BBH assembles diverse tasks drawing problems from linguistics, child development, maths, and common-sense reasoning, a.o. 
%among others.

\paragraph{Interactive games}
We take  \texttt{clembench}~\cite{chalamalasetti-etal-2023-clembench} as a benchmark to assess LLMs' gameplay ability in dialogue games.  We consider the games 1)~\textit{Taboo}, 2)~\textit{Wordle (base)} and the two variants \textit{Wordle (Clue)} and \textit{Wordle (Critic)}, 3)~\textit{Reference Game}, 4)~\textit{Image Game}, and 5)~\textit{Private/Shared}. Game descriptions and examples are provided in Appendices~\ref{appendix:clembench} and \ref{appendix:clembench-samples}.

\begin{figure}[h!]
    \centering
    \begin{tabular}{l}
    \includegraphics[width=0.47\textwidth]{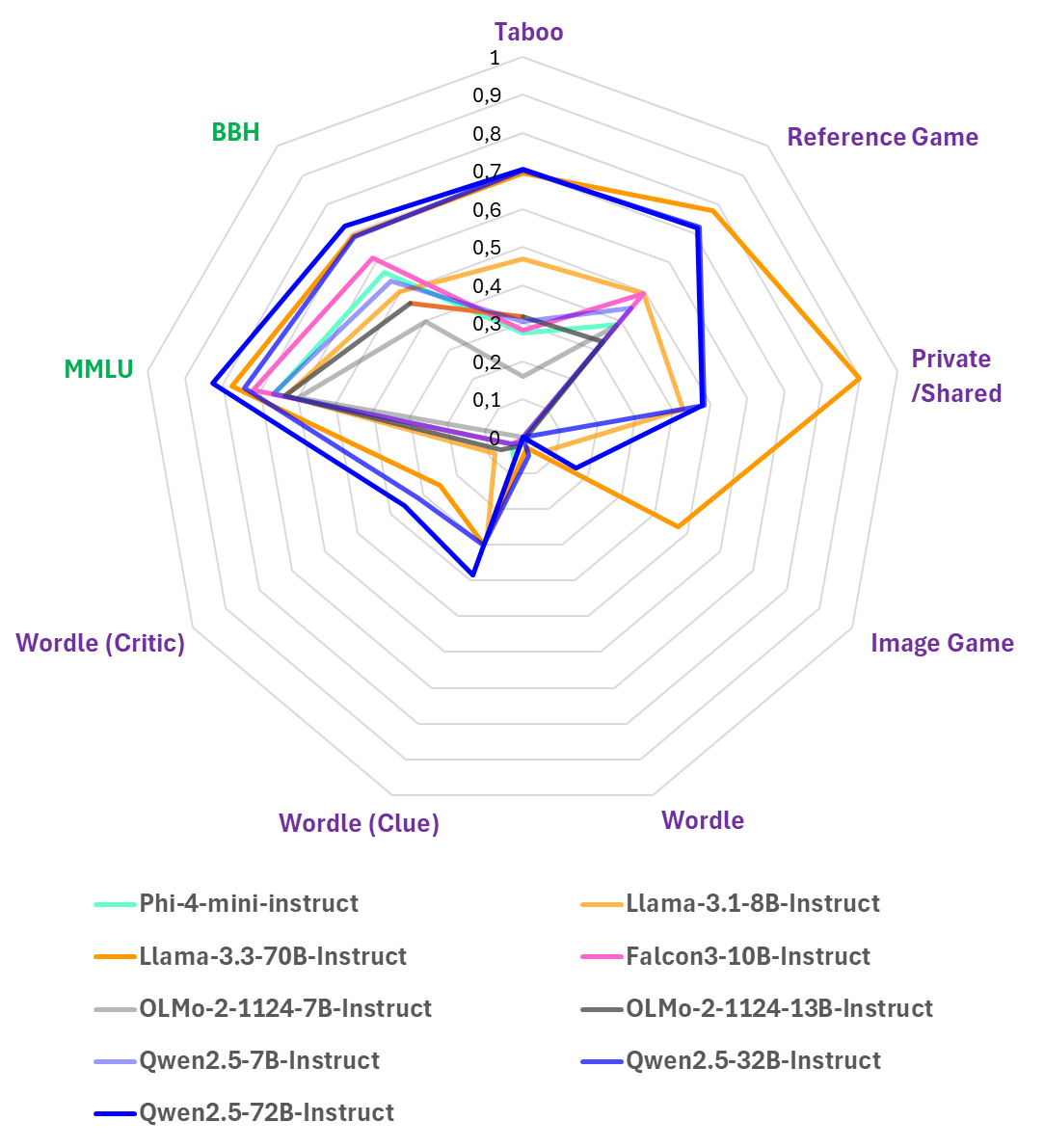}
    \end{tabular}
    \caption{Accuracy for all models on  \textcolor{massive_benchmarks}{\bf standard benchmarks}
     and \textcolor{interactive_games}{\bf interactive games}. See Table~\ref{tab:tab-all} in App.~\ref{sec:sup-plots}.}\label{fig:accuracy-ifevl-mmlu-taboo}
    \vspace{-0.4em}
\end{figure}

\begin{figure*}[h!]
    \centering
\begin{tabular}{c|c}
 \includegraphics[width=0.42\textwidth]{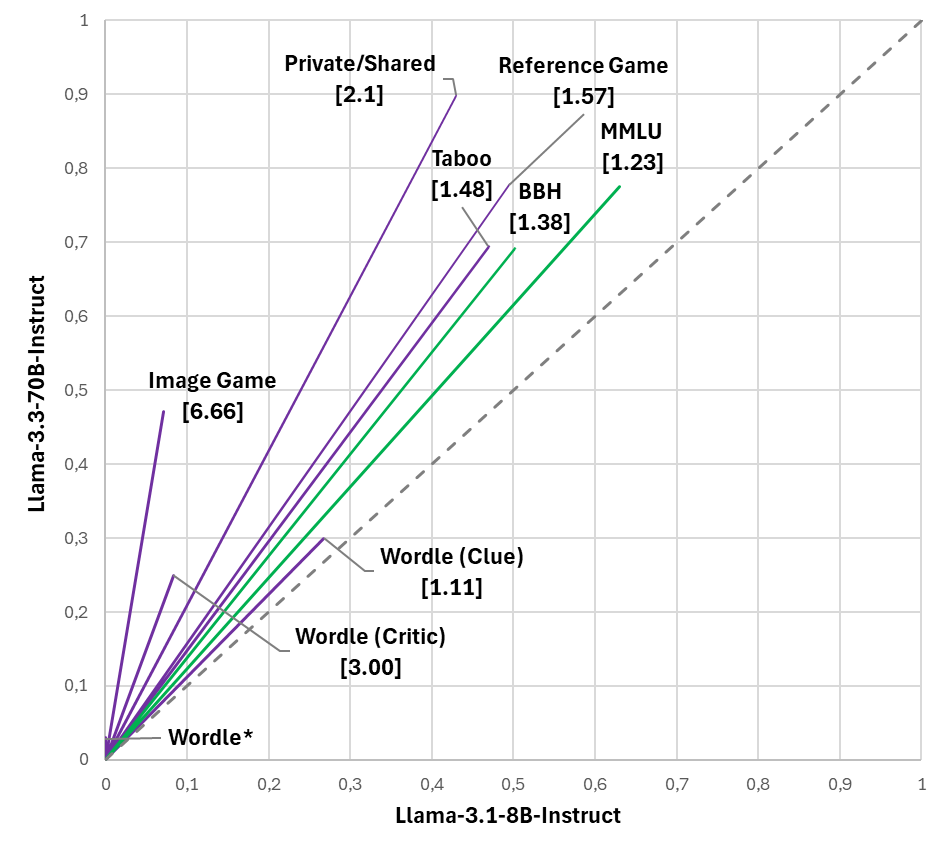 }
 \hspace{0.1cm}
 &

 \includegraphics[width=0.51\textwidth]{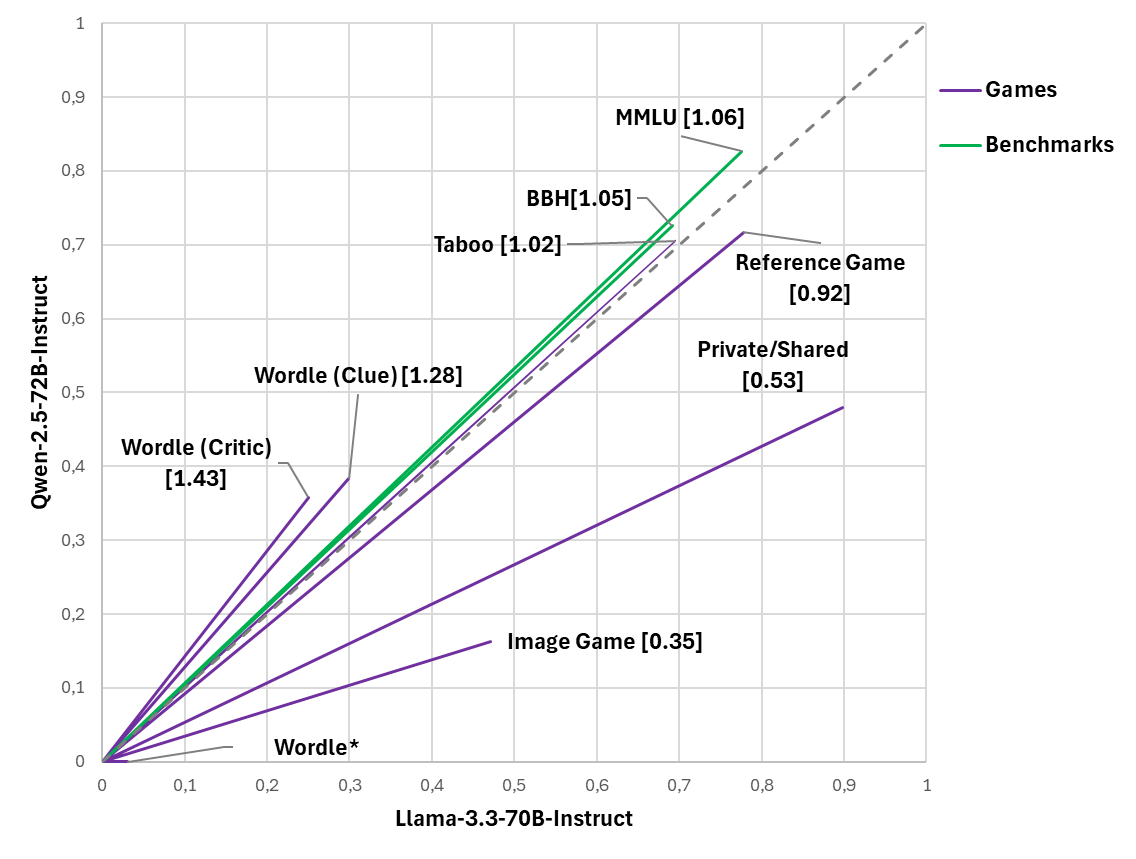} \\
 \includegraphics[width=0.42\textwidth]{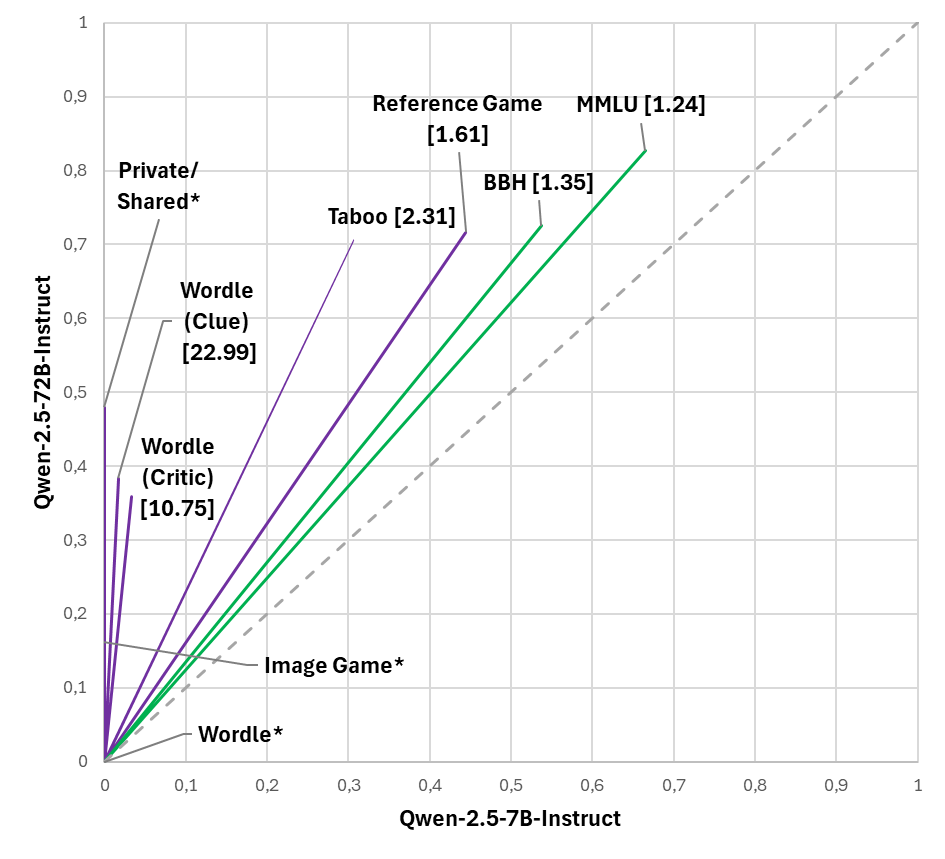} &
   \hspace{-1.35cm}
 \includegraphics[width=0.435\textwidth]{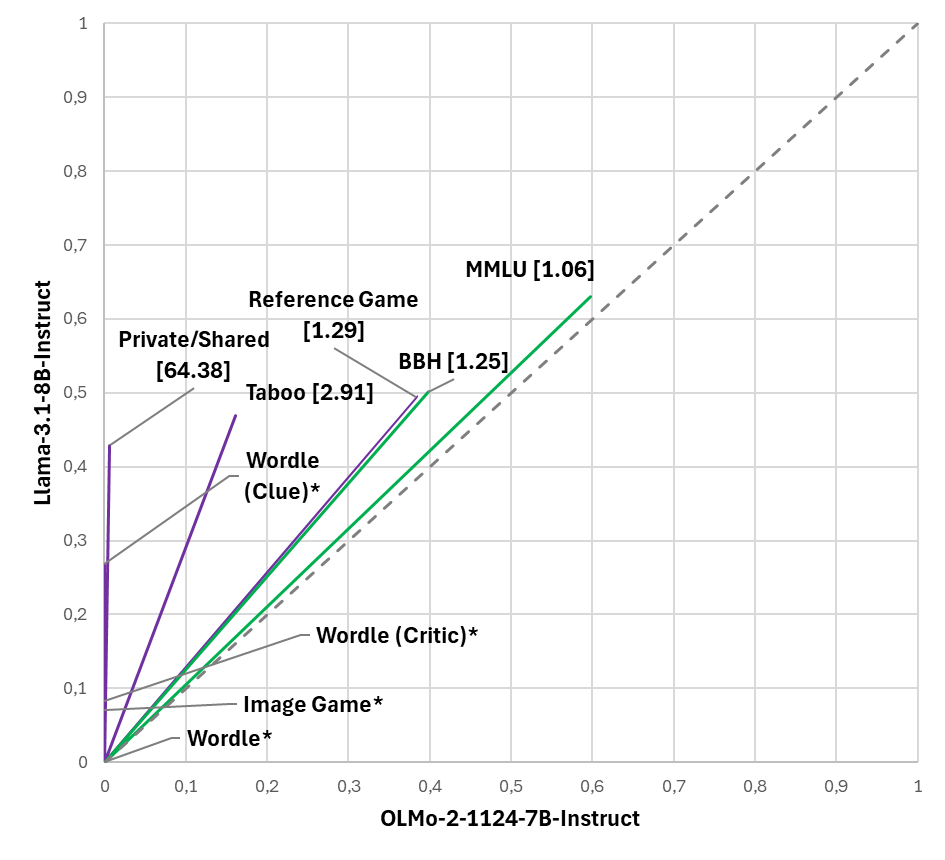} \\  
  \includegraphics[width=0.42\textwidth]{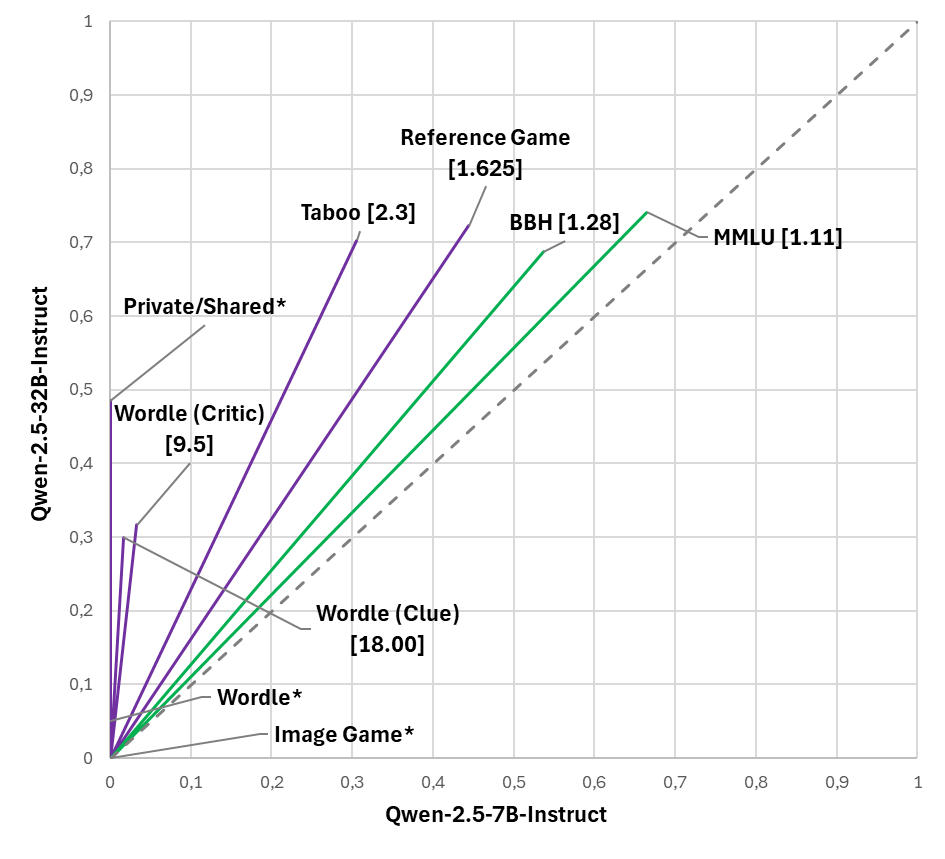} & 
  \hspace{-1.35cm}
  \includegraphics[width=0.435\textwidth]{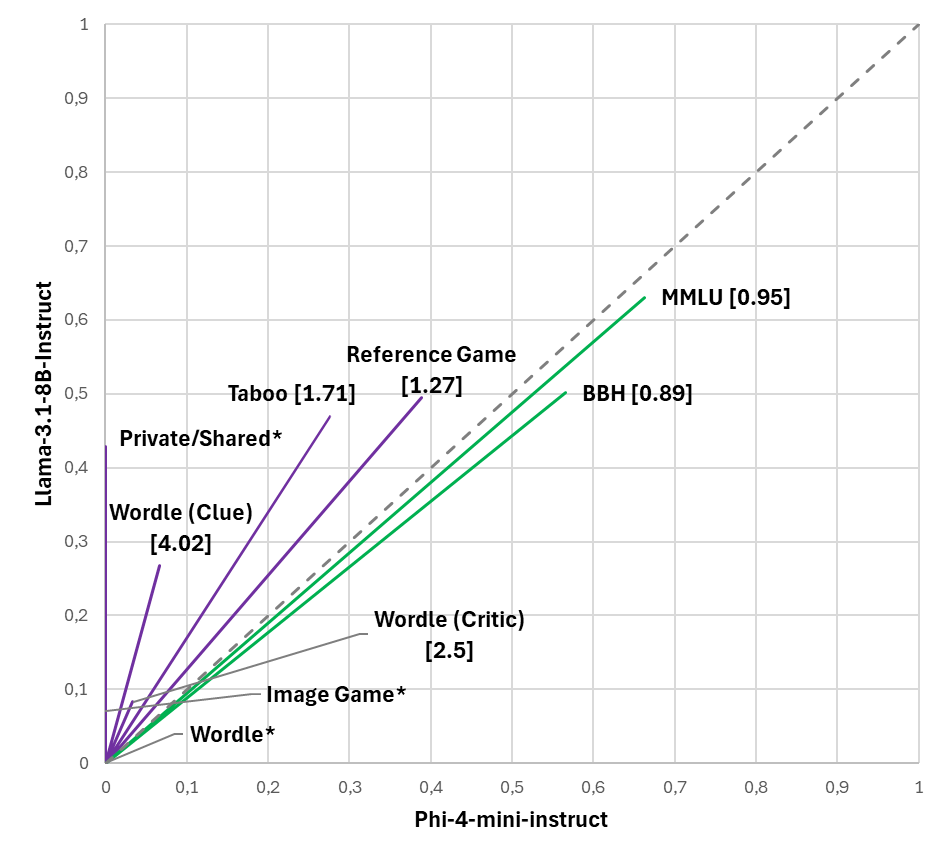} \\
 \end{tabular}
   \caption{Comparing datasets in their power to discriminate between models of different size but same family (left) and of different families but similarly large (right). The number next to the benchmark's name indicates the ratio of performance between the two models. The asterisk `*' next to \textit{Wordle} indicates that the ratio is undefined.}\label{fig:dataset-discriminative}
\vspace{-0.8em}
\end{figure*}

\paragraph{Results}
Figure~\ref{fig:accuracy-ifevl-mmlu-taboo} 
shows models' performance on the standard benchmarks and interactive games. %As mentioned in %Section~\ref{sec:models}, all models %we test are reasonably able to follow instructions as measured by IFEval. 
While the OLMo-2 models are more inconsistent across different model sizes, all the other models exhibit the expected pattern of showcasing better performance on both benchmarks and interactive games as the parameter count increases. At the same time, we observe that most of the interactive games highlight the benefits of larger model sizes much more strongly. This can more easily be appreciated in Figure~\ref{fig:dataset-discriminative} (left). In this visualisation, the further away a benchmark is from the diagonal, the more affected performance is by model size. While \textit{Wordle} is an extremely challenging game for any model, scaling up the number of parameters appears to be fundamental to succeed at \textit{Private/Shared}, \textit{Image Game}, and \textit{Reference Game}---much more than what can be observed for MMLU and BBH.\looseness-1 
    
Is size however all we need? 
Figure~\ref{fig:dataset-discriminative} (right) shows that reference-based benchmarks do not substantially distinguish between %large 
models of comparable size: Scaling on the number of parameters results in performance boosts across model families. Hence, arguably, reference-based benchmarks test for abilities that can be expressed within parametric knowledge. Given that such benchmarks are currently the standard LLM evaluation paradigm, it is not surprising that scaling is high on the agenda of model developers.
In contrast, interactive games seem to provide a different picture: models with comparable parametric capacity perform very differently on \textit{Image Game}, \textit{Private/Shared}, and \textit{Wordle (Clue/Critic)}. Similar trends can be observed in other model comparisons (see Appendix~\ref{sec:sup-plots}). This result supports the hypothesis that size is not all there is behind the potential of LLMs to learn inferential strategies for effective language use in interaction. 
%(See further details in App.~\ref{sec:sup-plots}.)

%====================================
\section{Cognitive Abilities}
\label{sec:eval_ca}
%====================================

% from RB in intro
%We call for the need to accompany these empirical studies with controlled experiments aimed to test the \textbf{cognitive abilities} that are conjectured to be behind the models' performance. Ideally, to carry out such analysis we should employ cognitive tests tailored for machine intelligence~\cite{hagen:machinepsy}. The need of developing an ``artificial cognition'' evaluation tool kit is made clear by~\cite{millire2024anthropocentric}, who discuss examples of auxiliary factors that could make human cognition test unsuitable for LLMs, yet they acknowledge the lack of LLM-tailored cognitive tests, and claim that an investigative process can and should start with cognitive tests designed for humans aiming at an  ``artificial cognition'' through a cyclic process. Following this literature, we propose a taxonomy of cognitive skills motivated by neurocognitive science and compile a list of existing evaluation datasets designed to assess each skill in isolation. 
%\rb{We leave to the future the revision of such a battery with LLM-tailored ones that might be inspired by studying the limitation of the cognitive tests currently at disposal and by triangulating LLMs with different evaluation methods.}

We now turn to targeted cognitive tests, which 
%---a complementary evaluation method that focuses 
assess specific cognitive abilities deemed essential for effective language use in real-world situations. We explore the use of such tests as a complementary evaluation method to  reference-based benchmarks and interactive games.

Ideally, such an investigation should be carried out using tests tailored to machine intelligence, which unfortunately 
have not yet been systematically developed  
%do not yet exist 
\cite{hagen:machinepsy}. To make progress in this direction, \citet{millire2024anthropocentric} advocate for the development of an ``artificial cognition evaluation toolkit'' through a cyclic process that takes cognitive tests designed for humans as the starting point. Following this literature, we present a taxonomy of abilities and a suite of existing evaluation datasets inspired by human cognition, with the aim to make 
%hope that our study will constitute 
a first step towards more adequate testing of artificial cognition. 

\subsection{Taxonomy and Datasets}

We present a taxonomy of cognitive abilities involved in human \textit{functional linguistic competence} \cite{maho:diss24}. 
%The taxonomy we present 
It is guided by neurocognitive research~\cite{ward:guide19}, and it separates capabilities into two distinct macro-categories known to recruit different brain networks: executive functions and socio-emotional skills. \textbf{Executive functions} are broadly defined as the complex processes by which we control and optimise our thoughts and behaviour \cite{baddeley:WM86}, divided into \emph{core} and \emph{higher-order} abilities. 
%, where the latter subsume the former. 
%for self-monitoring and error corrections we only found datasets based on human evaluation~\cite{tyen-etal-2024-llms,lin-etal-2024-criticbench}}.
%To evaluate higher-order abilities, we consider the following datasets:  \textsc{Natural Plan}~\cite{zheng2024naturalplanbenchmarkingllms}, to evaluate ``planning'', \textsc{CLadder}~\cite{jin2023cladder} for causal reasoning, LogiQA 2.0 ~\cite{logiqa} for logical reasoning, WinoGrande ~\cite{winogrande} which evaluates ``commonsense reasoning". To evaluate core abilities, we include the WCST and the LNT test ~\citep{kennedy2024cognitive} for ``cognitive flexibility'' and a \textit{n-back} task ~\citep{workingmemory} for ``working memory''.
\textbf{Socio-emotional skills} are abilities necessary to interact adaptively with other individuals \cite{higgins:ARP87}, including the ability to recognize their emotional and cognitive states. 
%See App.~\ref{subsec:appendix-taxonomy} for detailed descriptions.

For each cognitive ability, we select an existing evaluation dataset designed to test it in isolation. 
%drawing inspiration from human cognitive assessments. 
Socio-emotional skills have only recently entered the evaluation landscape in NLP, and they have done so with a forceful presence: remarkably, small benchmarks already exist for almost all 
of the abilities in this category.
%We discard datasets that require manual evaluation from the analysis. 
Tables~\ref{tab:exe-func} and~\ref{tab:soc} list the abilities in the taxonomy and the datasets we use to evaluate them.\footnote{We found no dataset to evaluate inhibitory control. The datasets we found for Emotion-regulation, Self-awareness~\cite{liu-etal-2024-interintent}, Empathy~\cite{chen-etal-2024-emotionqueen} and Social Problem-solving~\cite{du:llmfdecision24} require human evaluation.} More details are available in App.~\ref{subsec:appendix-taxonomy} and \ref{appendix-cognitivedatasets}.

\begin{table}[t] 
\resizebox{\columnwidth}{!}{
\begin{tabular}{@{}lll@{}} 
\toprule
& \textbf{Cognitive Ability} & \textbf{Benchmark} \\ 
\midrule
\multirow{3}{*}{\rotatebox{90}{Core}} &
Cognitive Flexibility  & \citet{kennedy2024cognitive} \\
& Working Memory  & \citet{workingmemory} \\
& Inhibitory Control  & -- \\\hline 
\multirow{4}{*}{\rotatebox{90}{HO}} &
Causal Reasoning & \citet{jin2023cladder} \\
& Common-sense Reasoning & \citet{winogrande} \\
& Logical Reasoning  & \citet{logiqa} \\
%Numerical Reasoning & -- \\ 
& Planning  & \citet{zheng2024naturalplanbenchmarkingllms} \\
%Self-Monitoring \& Correction & -- \\
\bottomrule
\end{tabular}
}
\caption{Core and Higher-Order Executive Functions.}
\label{tab:exe-func}
\vspace{-0.2em}
\end{table}

\begin{table}[t] \small \centering
\begin{tabular}{@{}ll@{}}\toprule
 \bf Cognitive Ability & \bf Benchmark \\ \midrule
Attribution and Judgement & \citet{gu:simpletom25} \\
Emotional Intelligence & \citet{paech2023eqbench}\\
Pragmatics & \citet{hu-etal-2023-fine} \\
Social Common-sense Reasoning & \citet{sap-etal-2019-social}\\
Theory of Mind & \citet{gu:simpletom25} \\
Emotion Regulation & -- \\
Self-Awareness & -- \\ 
Empathy & -- \\ 
Social Problem-Solving & -- \\
\bottomrule
\end{tabular}
\caption{Social and Emotional Skills.}
\label{tab:soc}
\vspace{-1em}
\end{table}

\subsection{Cognitive Ability Analysis}
\begin{figure}[h!]
    \centering
 \includegraphics[width=\columnwidth]{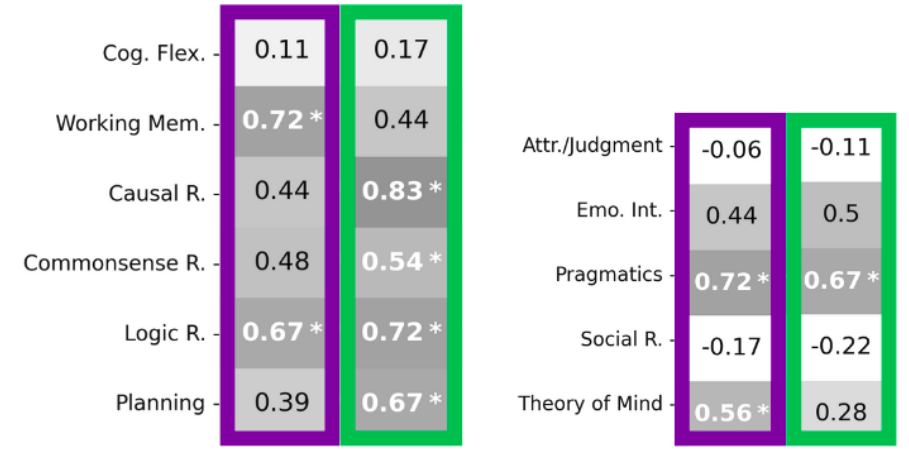}
 \caption{Correlation of cognitive abilities with performance on \textcolor{interactive_games}{\bf games} and \textcolor{massive_benchmarks}{\bf benchmarks} (* = $p< 0.05$).}
 \label{fig:cogabilities}
 \vspace{-0.9em}
\end{figure}

%====================================
Equipped with our taxonomy and associated cognitive tests, we aim to shed some light on the cognitive abilities involved in interactive games and general benchmarks. 
%Besides direct model comparisons, this analysis allows us to explore whether there is any correlation between models' cognitive abilities and their performance on static or interactive benchmarks. 
Figure~\ref{fig:cogabilities} reports Kendall's $\tau$ correlation coefficients.
The analysis reveals that performance both on static and interactive evaluation correlates with performance on tests measuring higher-order reasoning abilities: while planning is more dominant in static problem-solving tasks, working memory seems to be beneficial for games. Among the social skills, pragmatics appears to be relevant for both static and interactive tests, while ToM correlates better with the latter.

These results suggest that interactive tests correlate more strongly with socio-emotional skills compared to static tests, despite this analysis remaining speculative in nature, as we still lack carefully curated cognitive tests specifically designed for LLMs. %\looseness-1
Expanded results reporting correlation coefficients for each game and benchmark considered are shown in Appendix~\ref{sec:sup-plots}, Figure \ref{fig:cogabilities_complete}. Specifically for games, 
in Appendix~\ref{appendix:clembench-samples} we provide a qualitative analysis where we aim to draw 
%a qualitative analysis of transcripts of game episodes played by models (provided in  Appendix~\ref{appendix:clembench-samples}) shows that it is possible to draw 
tentative connections between capabilities and gameplay. 
%, although not for each game and  each capability. 
Remarkably, such connections appear to be clearer for executive functions compared to socio-emotional skills.\looseness-1

%how models' performance on MMLU, BBH, and interactive games correlates with the spectrum of their cognitive abilities. 
%The models' performance on clembench Reference game seems to correlate with their working memory ability (higher the former, higher the latter) --Kendall's Tau correlation (0.69); instead, for instance, social intelligence does not look to have any role in Clembench Taboo -- as one would expect given the nature of this game. Our analysis relies on the set of cognitive ability tests the community has produced so far; clearly work needs to be done on them for such analyses to become more reliable.

%\begin{figure}[h!]
    %\centering
%\begin{tabular}{cc}
% \includegraphics[width=0.45\textwidth]{latex/images/geometric-fig/cognitive_abilities_all.png} \end{tabular}
% \caption{Cognitive Abilities Spectrum} \label{fig:cognitive-abilities}
%\end{figure}

%\begin{figure*}[h!]
 %   \centering
%\begin{tabular}{ccc}
 % \includegraphics[width=0.35\textwidth]{latex/images/geometric-fig/cognitive_abilities_all.png} &
 %\includegraphics[width=0.45\textwidth]{latex/images/correlation_final-kendall/scatter-plotts/referencegame_wm.png} & \includegraphics[width=0.4\textwidth]{latex/images/correlation_final-kendall/scatter-plotts/tamboo_social.png}
 % \end{tabular}
  % \caption{Cognitive Abilities Spectrum (left) and Correlation with Cognitive Abilities: high (middle) vs.\ low (right) correlation}  \label{fig:cogabilities}
 %\end{figure*}

%\todo[inline]{
%Which cognitive abilities correlate with performance gain in multi-task static benchmarks vs interactive games. Figure~\ref{fig:correlation} and Figure~\ref{fig:scatterplots}
%}

%\begin{figure*}[h!]
 %\caption{Correlation with Cognitive Abilities}
  %  \centering
 %\includegraphics[width=1\textwidth]{latex/images/correlation_final-kendall/correlation_with_ca.png}
%\end{figure*}
 %   \label{fig:correlation}

%====================================
\section{Related Work}
\label{sec:related}
%====================================

A few recent papers have evaluated the abilities of LLMs beyond standard benchmarks. For example, \citet{waldis:holmes24} propose a framework to evaluate models' linguistic competence (in \citealt{chomsky1965}'s sense) using probing classifiers to disentangle knowledge of, e.g., syntax and semantics from other abilities such as instruction following. 
\citet{lu-etal-2024-emergent} posit that some perceived functional linguistic abilities (e.g., related to emotional intelligence) may be the result of in-context learning, model memory, and formal linguistic knowledge,  whereas \citet{ma-etal-2023-towards-holistic} propose an evaluation protocol for assessing machine Theory of Mind based on a taxonomy of existing benchmarks driven by psychology research. 

In contrast, the present work includes a uniquely comprehensive selection of evaluation tasks, encompassing standard benchmarks, interactive games, and cognitive ability assessments. To our knowledge, this diverse range of evaluations has not been jointly tested in prior work. Moreover, we systematically compare their discriminative power and analyse their interrelations---an approach that, to our knowledge, is unprecedented.

%====================================
\section{Conclusion}
%====================================

In this paper, we have examined three LLM evaluation paradigms. Our analyses revealed that interactive games have superior model discrimination power over standard reference-based benchmarks: Models with comparable parametric capacity perform similarly on standard benchmarks, but very differently on games. This suggests that scaling is not sufficient to learn strategies for effective language use in interaction. 

We have further argued for the use of targeted cognitive tests as a complementary  method to triangulate LLM progress. 
%Our results show the superior discriminating power of interactive games over single-turn general-purpose reference-based benchmarks.
%Crucially, we argue that in order to claim that LLMs have emerging abilities, measuring performance on general benchmarks or interactive games is not sufficient per se, but should rather be %accompanied by correlation analysis triangulated with controlled tests designed to evaluate such abilities. Furthermore, we highlight the potential value of carefully designed controlled benchmarks inspired by human cognitive ability assessment as a good means for such correlation analyses. 
While each cognitive ability test alone does not get us very far in the quest for robust LLM evaluation, we contend that this type of evaluation paradigm has the potential to enhance our understanding of what fundamental abilities LLMs must develop to be able to function effectively as language agents, where multiple skills may be required and possibly interact. In our study, we have drawn inspiration from human cognitive tests as a first step in this direction. We hope to encourage the future development of targeted tests tailored to artificial cognition and
% Nevertheless, as as emphasised by \citet{millire2024anthropocentric}, caution should be exerted before drawing conclusions about LLMs' abilities from tests designed for humans. New carefully designed behavioural experiments for LLMs should be proposed, 
 supplemented with mechanistic analyses.

 %Hence, we call for future work for the creation  of new interactive benchmarks which can discriminate well between models and which correlate strongly with underlying cognitive abilities. 

%\tb{The longstanding link between AI and cognitive science has recently been strongly reinforced in several papers which describe it as a new emerging field, Machine Psychology\cite{hagen:machinepsy} or AI Psychology\cite{ivan:howt25}, the importance of profiting of the evaluation toolkit from developmental psychology has been advocated by~\citet{FRANK2023990}.}

\section*{Limitations}
The evaluation methods we have study in this papers have been applied 
%Our evaluation prompts models to provide direct answers 
without employing chain-of-thought (CoT) reasoning or similar elicitation techniques. 
While different elicitation strategies may enhance question-answering, interactive, and cognitive abilities in different ways \citep{shunyu-etal-2023-tree,hao-etal-2023-reasoning,li-etal-2024-emotional}, we opted for an approach that remains agnostic to specific prompting methods and datasets. 
This ensures a consistent basis for comparison across models, though future work could explore how alternative prompting strategies influence performance across the three evaluation paradigms. 

The set of models we evaluated is of course limited in scope, although representative of a wide range of open-weight model families and sizes. 
%We could have included many more models to have a more comprehensive analysis, 
In particular, we did not evaluate  any close-weight proprietary models, as we wanted to ensure full reproducibility of our experiments. %, and therefore we wouldn't want to risk evaluating models that would be deprecated in the future.
%We subscribe to the view described here: https://hackingsemantics.xyz/2023/closed-baselines/.
 
Regarding the cognitive ability assessments, we have used currently available datasets; such resources have started to be compiled only very recently, hence there is no guarantee that in their current state they truly assess the intended cognitive abilities in LLMs. 
%tests we used may not guarantee to evaluate the intended abilities in LLMs. 
Nevertheless, they help in establishing a starting point and encouraging further progress in this direction.\looseness-1 

Finally, our evaluation is only on English data, hence we cannot claim the results hold for any language. Yet, we believe the main take-home message of the paper, viz.\ using different evaluation methods which include static and interactive tasks as well as controlled cognitive experiments, holds independently of the language analysed. Whether this is indeed the case, remains an interesting open question for future research. 
%It will be indeed interesting to see whether different language elicit different correlations between controlled experiments and static/interactive tasks.

%\rf{add point on the fact that the cog abilities datasets may not be perfect to evaluate the intended abilities}

\section*{Acknowledgments}
This work emerged from discussions at a workshop organised by Raquel Fernández and Sandro Pezzelle at MFO, the Oberwolfach Research
Institute for Mathematics in the German Black Forest, on behalf of the ELLIS NLP programme. The event was funded by the state of Baden Württemberg (Germany) and organised in collaboration with the ELLIS Institute Tübingen and
the Max Planck Institute for Intelligent Systems.
Alessandro Suglia acknowledges the use of the Heriot-Watt University high-performance computing facility (DMOG) and associated support services in the completion of this work. 
Alessandro Suglia, Raffaella Bernardi, and David Schlangen acknowledge ISCRA for awarding this project access to the LEONARDO supercomputer, owned by the EuroHPC Joint Undertaking, hosted by CINECA (Italy). 
We furthermore acknowledge our funders.
Mario Giulianelli was supported in part by an ETH Zurich Postdoctoral Fellowship. 
David Schlangen acknowledges support by the Deutsche Forschungsgemeinschaft (DFG, German Research Foundation) – 423217434 (“RECOLAGE”) grant, and support from University of Potsdam (President's Fund). Raquel Fern\'andez was supported by the European Research Council (ERC Consolidator Grant DREAM 819455).
This research has been partially supported and sponsored by the Province of Bolzano and the EU through the project Artificial Intelligence Laboratory (AI-Lab)
ERDF 2021-2027 EFRE1047, CUP: I53C23001670009 (01/01/2024 - 31/12/2026).

\bibliography{biblio}

\appendix

\section*{Appendix}

\section{Models}
\label{sec:appendix-models}
The Phi-4 series~\cite{phi4} includes a model with 3.8 billion parameters. Despite its compact size, it has been reported to surpass larger models on a wide selection of tasks including reasoning, general-purpose and coding.

The Olmo-2-1124 series~\cite{olmo20242olmo2furious} includes models with 7 billion and 13 billion parameters (\texttt{OLMo-2-1124-*-Instruct}). Both models are designed for a variety of tasks, including chat, mathematics, and reasoning. They have undergone supervised fine-tuning on the Tülu 3 dataset and further training using DPO techniques.

The Qwen2.5 series~\cite{qwen2, qwen2.5} includes models with 7B, 32B, and 72B parameters (\texttt{Qwen2.5-*-Instruct}). They are multilingual, supporting over 29 languages, and excel in coding, mathematics, and instruction following.

The Llama-3 series~\cite{dubey2024llama3herdmodels} includes models with 8B (\texttt{Llama3.1-8B-Instruct}) and 70B parameters (\texttt{Llama3.3-70B-Instruct}). These models are optimized for multilingual dialogue and support various languages. They use an optimized transformer architecture and are fine-tuned for instruction following.

The Falcon3 series~\cite{falcon3} includes a model with 10 billion parameters. It achieves state-of-the-art results in reasoning, language understanding, instruction following, code, and mathematics tasks. It supports four languages (English, French, Spanish, Portuguese) and a context length of up to 32K.

\paragraph{Model Implementations}
All the models used in this study have been made available by Huggingface, and have been accessed through the \textit{transformers} \cite{wolf-etal-2020-transformers} library.
For text generation, we have been applying the default chat template specified by the models' Hugginface implementations.

% \subsubsection*{Taxonomy of Cognitive Abilities}
\section{Taxonomy of Cognitive Abilities}
\label{subsec:appendix-taxonomy}

\noindent
\textsc{Executive Functions:}
\begin{itemize}[leftmargin=10pt]
\item \textit{Core abilities}
\begin{description}
\item[Working Memory:] Hold and manipulate information in mind over short periods;
\item[Inhibitory Control:] Suppress automatic, inappropriate, or impulsive responses and resist distractions;
\item[Cognitive Flexibility:] Adapt to new situations, switch between tasks, and think about multiple concepts simultaneously;
%\item[Self-monitoring and Error correction:] Assess one's performance and behavior to ensure they align with desired outcomes, or correct otherwise.
\end{description}

\item \textit{Higher-order abilities}
\begin{description}

\item[Planning:] Set goals, develop steps to achieve them, and anticipate potential obstacles;
\item[Causal Reasoning:] Understand cause-and-effect relationships;
\item[Logical Reasoning]: Deductive and inductive reasoning;
%\item[Numerical Reasoning]: Quantitative reasoning and arithmetic problem-solving.
\item[Common-sense Reasoning:] Apply general common knowledge to everyday scenarios, including understanding basic physical properties, such as gravity, solidity, and object interaction;
%\item[Self-monitoring and Error correction:] Assess one's performance and behavior to ensure they align with desired outcomes, or correct otherwise. 
\end{description}
\end{itemize}

\noindent
\textsc{Socio-emotional skills:}
\begin{description}
\item[Social Common-sense Reasoning:] Understand social norms and expectations;
\item[Social Problem-Solving:] Analyze social situations, generate solutions, and make decisions that foster positive interactions;
\item[Emotional Intelligence:]  Recognize, interpret, and 
manage one’s own and others’ emotions.
\item[Emotion Regulation:] Manage and modify one's emotional responses in appropriate ways;
\item[Self-Awareness:] Recognize and understand one's own emotions, thoughts, and behaviors;%subgroup of EMI?
\item[Empathy:] Share and understand the feelings of others, both emotionally and cognitively;
\item[Theory of Mind:] Understand that others have thoughts, beliefs, desires, and intentions different from one's own;
\item[Attribution and Judgment:] Interpret the causes of others' behavior, distinguishing between intentional and unintentional actions.
\item[Pragmatics:] Aspects of communication that go beyond formal language competence: considering communicative intentions, the communicative context of the utterance, shared knowledge between speakers, manners, social and cultural norms.
 \end{description}

\section{Interactive Games}\label{appendix:clembench}

We leverage \textit{clembench}~\cite{chalamalasetti-etal-2023-clembench}, a benchmark that assesses models' gameplay ability in well-defined dialogue games such as:
\begin{description}
    \item[Taboo:] A game where one player tries to get the other player to guess a word without using certain `taboo' words in their clues; 
    \item[Wordle:] A game where one player thinks of a 5-letter word and the other player tries to guess it. At each turn, the guesser receives feedback regarding whether each letter of the tentative word is present in the target word, and whether it is also in the same position in such word;
    \item[Wordle (Clue):] A variant of Wordle where the guesser is given more clue to help them guess the target word; 
    \item[Wordle (Critic):] A variant of Wordle where the guesser's suggestions are evaluated by a third player, who provides feedback on their accuracy; 
    \item[Reference Game:] A game where one player is presented with three textual grids and is tasked to make another player---who is also presented with the same three grids, but potentially in a different order---identify the one it has in mind based on its textual description;
    \item[Private/Shared:] A game where a customer agent goes through a form with a service agent and, after each turn, a third agent,\footnote{
    In clembench, all interactions are mediated by a ``Game Master''. This agent plays a particularly active role in \textit{Private/Shared}.
} probes the customer on what they believe that the service agent already knows;
    \item[Image Game:] A game where given a certain grid structure similar to those in \textit{Reference Game}, a player has to instruct a second one (who can't see the it) about how to reconstruct it step by step.
\end{description}

\section{Interactive Games: Samples}
\label{appendix:clembench-samples}
In this section we provide samples for the interactive games, and discuss their nature and links with cognitive abilities, as they are shown by the correlation analysis displayed in Figure \ref{fig:cogabilities_complete}. We provide samples deriving from our our evaluations however, as a reference, the clembench benchmark's maintainers made available a repository of transcripts for the conversations obtained by evaluating several models on the games.\footnote{\url{https://github.com/clembench/clembench-runs}}
\subsection{Taboo}
Figure \ref{taboo_transcript} shows an episode of Taboo played by Llama-3.1-70B-Instruct. In this case, the word to guess is \textit{clear}, and the Describing player (on the left) is instructed to generate a description for the other player without using the taboo words \textit{crystal}, \textit{obvious} and \textit{confident}. The description is then delivered to the guesser who guesses the word \textit{glass} at first. Since this was not the correct word, the game proceeds for another round, when the guesser gets the right word.

As shown in Figure \ref{fig:cogabilities_complete}, this game presents a higher correlation with working memory, common-sense and logical reasoning among the executive functions, as well as pragmatics and Theory of Mind among the socio-emotional skills.

We can observe from the example interaction that the correlation results are quite compatible with the abilities required for being successful at the game. Common-sense reasoning is exercised for guessing the appropriate word from the descriptions, and for the generation of descriptions themselves, along with Theory of Mind which is necessary for tuning the descriptions with the guesser's knowledge that can be inferred from its answers.
Logical reasoning can be instead exercised along with common-sense reasoning for guessing.
Working memory can be used to avoid repeating the same answer and being able to remember useful information from the conversation, while the role of pragmatics remains less clear.
\subsection{Wordle (and its variants)}
Figure \ref{wordle_transcript} shows an episode of Wordle. The model here is asked to guess the word \textit{whine}. At each turn, the model receives letter-level feedback which has to exploit in order to get to the solution.
We have evaluated models also on the variants providing additional clue (Figure \ref{wordleclue_transcripts}) and the support of a critic model judging every guess (Figure \ref{wordle_critic_transcript}).
While correlation has not been measured on the classic Wordle variant due to the excessively low performance of models, we can interestingly observe differences in the correlation between the two variants and the cognitive capabilities. 
Results suggest that the presence of an external model stimulates the use of social capabilities.
\subsection{Image Game}
Figure \ref{imagegame_transcripts} shows a match of Image Game, where a \textit{describer} player has to describe a certain grid step by step to a second player which has to draw it by following the instruction. The describer has then to decide when the grid is complete. Given the high difficulty of the game for most of the models in this study, inferring links with cognitive capabilities appears complicated.
\subsection{Reference Game}
In Figure \ref{reference_transcript} is displayed an episode of Reference Game where a \textit{describer} model has to describe a given grid with a empty cross-like shape. The other player has to guess the grid by following the description.

In our study, this task has been shown to correlate with Planning more than other cognitive capabilities. This connection appears reasonable: in order to build the best description a model may either describe the grid cell by cell, or it may look at the task as a constrain resolution problem where it has to resolve conflicts in order to provide the the simplest description that does not overlap with that of other grids. In this case, it is possible to observe analogies with schedule planning tasks such as those present in our planning benchmarks. 
In the case of other executive functions, working memory may be involved to remember the information regarding the grids within the context, and logical reasoning for guessing the appropriate grid.

Less clear are the correlations with social functions.
\subsection{Private/Shared}
Private/Shared is a game where two parallel conversations occur at the same time. Figure \ref{privateshared_transcripts} shows the main conversation of an episode simulating a travel-agency interaction. Among the games taken into consideration, it is the most conversational task. 
At each turn, in a parallel conversation the player is questioned regarding its understanding of the state (private or public) of information.

The low correlation of this task with Theory of Mind is quite surprising and would be worth further investigations. On the contrary, it appears understandable its correlation with working memory superior to that of any other tasks, as conversations are generally longer and require more explicit recalling of information shared in prior messages.

% \subsubsection*{Benchmarks used to evaluate the Cognitive Abilities}
\section{Benchmarks for Cognitive Abilities}
\label{appendix-cognitivedatasets}
% \begin{itemize}
%     \item[WorkingMemory:] 
%     \item[LLM-Cognitive-Flexibility:]
%     \item[LogiQA 2.0:]
%     \item[CLadder:]
%     \item[WinoGrande:]
%     \item[NATURAL PLAN:]
%     \item[EQBench:]
%     \item[LM-Pragmatics:]
%     \item[SocialIQA:]
%     \item[SimpleToM:]
% \end{itemize}

\begin{description}
    \item[Working Memory] ~\cite{workingmemory} (referred as \textit{WM} in this work)  is a set of verbal and spatial n-back tasks presented with three levels of difficulties from $n=1$ to $n=3$. The model has to identify whether the current stimulus (a letter in a string or a spatial location in a grid) is the same as the stimulus shown $n$ messages back or not. Size: 50 blocks of 24 instances each, total: 1200;

    \item[Cognitive Flexibility] ~\cite{kennedy2024cognitive} (referred as \textit{LLM-Cognitive-Flexibility} in this work) aims to test to what degree LLMs can rapidly switch tasks within a single context window. To this end, it employs two neuropsychological tests, the Wisconsin Card Sorting Test (WCST)~\cite{grant:wcst48} and the Letter-Number Test (LNT)~\cite{roger:costs93} commonly used to measure cognitive flexibility in humans. Size: 8 runs by 25 instances each. Total: 200;

    \item[Logical Reasoning] \textit{LogiQA 2.0}~\cite{logiqa} This dataset evaluates logical reasoning using the same data both in NLI and Machine Reading Comprehension format (text, question, multiple-choice) for each of the following (deductive) reasoning types: categorical, sufficient condition, necessary condition, disjunctive, conjunctive reasoning. Size: 1572 data points;

    \item[Causal Reasoning] \textsc{CLadder}~\cite{jin2023cladder} focuses on formal causal reasoning (causal inference), as opposed to common-sense causal reasoning. The dataset is constructed from formal logic-based templates that are then verbalised into natural language as binary questions. Size: 10112 data points;

    \item[Common-sense Reasoning] \textit{WinoGrande}~\cite{winogrande} A large-scale dataset of common-sense reasoning problems consisting of pairs of nearly identical questions with two answer choices (as in the original Winograd Schema Challenge~\cite{morgenstern:wsc15}, from which this dataset takes inspiration but aims to make larger and more complex. Size: 1267 data points;

    \item[Planning] \textsc{Natural Plan}~\cite{zheng2024naturalplanbenchmarkingllms} is a realistic planning benchmark consisting of three tasks expressed in natural language: Trip Planning, Meeting Planning and Calendar Scheduling. Models are given a situation and a problem to solve (e.g. find a trip plan that satisfies some constraints given the situation described). Each task contains problems of different levels of complexity based on the number of cities, people or days involved. The problems are often based on numerical reasoning too. We evaluate models on the Trip Planning and the Calendar Scheduling tasks. Size: 2600 data points;

    \item[Emotional Intelligence] \textit{EQ-Bench}~\cite{paech2023eqbench} the model is given an emotionally charged short dialogue (generated by GPT-4) and has to score the four possible emotions felt by a given character. Scores are compared against a reference score.  Size: 171 data points;

    \item[Pragmatics] \cite{hu-etal-2023-fine} (referred as \textit{LM-Pragmatics} in this work) is a benchmark evaluating LLMs' understanding of seven pragmatics phenomena: deceit, indirect speech, irony, maxims, metaphor, humour, and coherence. Scenarios are grounded into social situations, requiring LLMs to interpret utterances. The task is designed as a multi-choice questionnaire with 2-5 questions based on the subtask. Size: 845 data points;

    \item[Social Common-sense] \textsc{Social IQa} \cite{sap-etal-2019-social} a dataset for evaluating social common-sense reasoning and emotional intelligence. Each sample includes a short scenario and three multiple-choice questions across six categories: intentions, reactions, descriptions, motivations, needs, and consequences. Transfer learning on this dataset has shown strong performance on other common-sense reasoning benchmarks. Size: 1954 data points;

    \item[Attribution and Judgment/Theory of Mind] \textit{SimpleToM}\cite{gu:simpletom25} contains concise, diverse stories each with questions that ask models to predict behavior ("Will Mary pay for the chips or report the mold?"), judgment ("Mary paid for the chips. Was that reasonable?") or mental states ("Is Mary likely to be aware that 'The can of Pringles has moldy chips in it.'? Yes or No?") The first two subtasks have been taken as a reference for the Attribution and Judgment cognitive ability, while the last as a reference for Theory of Mind. Size: 3450 data points.
\end{description}

%\section{Dimensions of the Cognitive Benchmarks}
%As a reference, we provide information regarding the dimensions of the selected cognitive benchmarks in terms of sample size:
%\begin{itemize}
%    \item \textit{WM}: (50 blocks of 24 instances each). Total: 1200;
  %  \item \textit{LLM-Cognitive-Flexibility}: 8 runs by 25 instances each. Total: 200;
   % \item \textit{LogiQA 2.0}: 1572;
    %\item \textsc{CLadder}: 10112;
    %\item \textit{WinoGrande}: 1267;
    %\item \textsc{Natural Plan}: 2600;
   % \item \textit{EQ-Bench}: 171;
   % \item \textit{LM-Pragmatics}: 845;
   %\item \textsc{Social IQa}: 1954;
 %   \item \textit{SimpleToM}: 3450.
%\end{itemize}

% \subsubsection*{Benchmark Implementations}
\section{Benchmark Implementations}
For the majority of the static benchmarks evaluated in this work we relied on the popular framework for the evaluation of LLMs \textit{LM Evaluation Harness}\footnote{\url{https://github.com/EleutherAI/LM Evaluation Harnessuation-harness}} (ver. 0.4.7), which already made available many of the selected benchmarks, and enabled a common interface for the implementation of most of the remaining ones.  

The benchmarks which were already present within the framework are: \textsc{Social IQa}~\cite{sap-etal-2019-social}, WinoGrande \cite{winogrande}, EQ-Bench \cite{paech2023eqbench}, LogiQA 2.0 \cite{logiqa}, MMLU \cite{hendryckstest2021}, BBH \cite{suzgun-etal-2023-challenging}. 
The benchmarks which have been implemented in the framework over the course of the study are: \textsc{CLadder} \cite{jin2023cladder}, LM-Pragmatics \cite{hu-etal-2023-fine}, SimpleToM \cite{gu:simpletom25}, \textsc{Natural Plan} \cite{zheng2024naturalplanbenchmarkingllms}, IFEval \cite{zhou2023instruction}.

As for the interactive games, we have used the implementation provided by version 1.6 of the \textit{clembench} \cite{chalamalasetti-etal-2023-clembench}. The remaining benchmarks (WM, LLM-Cognitive-Flexibility) have been implemented outside of the framework, as \textit{LM Evaluation Harness} did not provide support for the multi-turn nature of the tasks. 

\subsection{Zero-shot and Few-shot Tasks}   
The majority of the tasks have been evaluated in a zero-shot setting with the exception of MMLU (5-shot), BBH (3-shot) (following common practices in model evaluation, e.g. in the Open Leaderboard 2\footnote{\url{https://huggingface.co/docs/leaderboards/open_llm_leaderboard/about}} for BBH) and \textsc{Natural Plan} (5-shot). In the case of \textsc{Natural Plan}, our models performed really poorly when evaluated in a zero-shot fashion---resulting in scores close to 0. Given that the task relies on the models producing answers in a strict format for parsing, we opted for using the 5-shot version provided by the benchmark's authors.

% Ale stopped reviewing here; to be continued.

\subsection{Metrics}
\paragraph{Evaluation}
For the evaluation of models, we followed the original works' implementations as well as associated metrics. However, it may be the case that for a certain benchmark more metrics were defined, or that the original work did not aggregate results across subtasks. For this reason, we report here the metrics we used for evaluating models.

In the case of \textit{Clembench games}, we computed performance by computing the ratio between the quality score (a number from 0 to 100) and the percentage of played games (a number between 0 and 1) divided by 100. 

In the case of \textit{IFEval}, following what was done in the Open Leaderboard 2, we averaged the results obtained on prompt-level and instruction-level strict accuracy.

As for \textit{EQ-Bench}, we computed the task-specific score as it was implemented in the \textit{LM Evaluation Harness}.

Regarding \textit{WM}, we only considered the subtask \textit{Verbal N-3}, and we computed the accuracy for the results obtained across the 50 trials defined in the original work.

In the case of \textit{LLM-Cognitive-Flexibility}, we ran each subtask 8 times with 25 trials each, and computed the average of the accuracy obtained in each run. In this case, the accuracy was computed only on the trials for which response parsing was successful. We then averaged the accuracy obtained on both subtasks to compute the final score.

In the case of \textsc{CLadder}, we followed the original work which treated the task as generative and probed for the presence of the substrings "yes"/"no" at the beginning of the model's answer.

In \textsc{Natural Plan}, the original work defined a rule-based procedure to parse specific data from the generated plan (e.g., dates). We reuse their parsing procedure and verify whether the expected elements are all present in the parsed plan.

For the remaining tasks (LogiQA 2.0, WinoGrande, LM-Pragmatics, \textsc{Social IQa}, MMLU, BBH, SimpleToM), we treated them as a multiple-choice question answering task that is evaluated based on the likelihood of the correct answer for the task.

In the case of BBH, the Open Leaderboard 2's evaluation code excludes three of the original tasks from the overall score's computing: \textit{dyck languages}, \textit{navigate} and \textit{word sorting}. The performances on these subtasks are therefore also ignored in the performance reported in this study.

In the case multiple subtasks were present (LM-Pragmatics, MMLU, BBH, \textsc{Natural Plan}, LLM-Cognitive-Flexibility), we computed the micro-average over the results achieved on each subtask. In the specific case of SimpleToM, since the subtasks were associated with two different Cognitive Abilities, we've aggregated the score of the subtasks \textit{behaviour} and \textit{judgment} into a single score (under Attribution and Judgment), and considered the \textit{mental state} subtask separately (under Theory of Mind).

\paragraph{Correlation}
For measuring the pair-wise correlation between benchmarks, we've computed the Kendall rank correlation coefficient (or Kendall's Tau) (Tau-b version). It measures rank correlation according to this formula: 

\[
\tau_b = \frac{P - Q}{\sqrt{(P + Q + T_x)(P + Q + T_y)}}
\]
where:
\begin{align*}
P &= \text{number of concordant pairs}, \\
Q &= \text{number of discordant pairs}, \\
T_x &= \text{tie correction for variable } X, \\
T_y &= \text{tie correction for variable } Y.
\end{align*}

This method was preferable over others given its robustness in case of few data points, as it was in our case.
We have also experimented with the Pearson correlation coefficient and observed that in the majority of the cases, the correlation patterns were similar, however with larger positive as well as negative correlations compared to Kendall.
We've relied on the implementation provided by the \textit{pingouin} Python package \cite{pingouin}.
\subsection{Generation Settings}
The tasks which required the models to generate text are: EQ-Bench, WM, BBH, IFEval, the clembench games, LLM-Cognitive-Flexibility, \textsc{Natural Plan}, \textsc{CLadder}.
With the exception of Working Memory and LLM-Cognitive-Flexibility, all tasks have been evaluated by applying a temperature of 0.
Following the original implementation, we have applied a temperature of 1 to WM and 0.7 to LLM-Cognitive-Flexibility. In these cases, however, the increased randomness caused by the higher temperature was mitigated by averaging the results obtained over multiple trials.

As for the other generation settings, we also have followed what was prescribed in the original works regarding the tokens for the termination of the generation, the maximum or minimum number of tokens. 
In the case of \textsc{Natural Plan}, the original work did not provide specific information regarding the settings they have adopted for the evaluation. Given the highly challenging nature of the task, we have set the minimum and maximum number of tokens to 90 and 350, respectively. This was derived based on the minimum and maximum number of tokens in the gold plans. 

\section{Limitations in the Evaluations}
In certain cases, results have not been computed on all the subtasks available for that benchmark.
In the case of \textsc{Natural Plan}, we have not considered results coming from the \textit{meeting} subtask, while for WM we have only considered those coming from the \textit{Verbal (Base) N-3} subtask. In the first case, the high amount of resources required for evaluating the task, especially for the larger models prevented us from doing so. As for the second, we've only considered the base version of the verbal subtask and excluded its variations as they would not provide meaningful information for this study.

\section{Computational Resources}
As a reference, we provide the time required for running through all the benchmarks for the Llama-3.1-8B-Instruct model on 1 A100 GPU with batch size set to 'auto' in the \textit{LM Evaluation Harness} (i.e. it automatically fits into the memory the maximum batch size possible for each task). For the Clembench games, LLM-Cognitive-Flexibility and WM, the batch size is 1. The time also includes that required for procedures performed by the \textit{LM Evaluation Harness} prior to the actual evaluation (relevant only for those datasets included in this framework) and for loading the model into the memory (all tasks).
\begin{itemize}[noitemsep]
    \item LLM-Cognitive-Flexibility: \char`~1:50 min
    \item LogiQA 2.0: \char`~5 min 
    \item \textsc{CLadder}: \char`~19:30 min
    \item WinoGrande: \char`~1 min
    \item \textsc{Natural Plan}: \char`~4:50 hours
    \item WM \char`~2:40 min
    \item EQ-Bench: \char`~3 min 
    \item LM-Pragmatics: \char`~6:30 min 
    \item \textsc{Social IQa}: \char`~1:30 min 
    \item SimpleToM: \char`~2:40 min 
    \item MMLU: \char`~14 min 
    \item BBH: \char`~24 min
    \item IFEval: \char`~1:19 min
    \item Taboo: \char`~3:30 min
    \item Reference Game: \char`~3:00 min
    \item Image Game: \char`~2.40 min
    \item Wordle: \char`~7:50 min
    \item Wordle (Critic): \char`~2:50 min
    \item Wordle (Clue): \char`~2:15 min
    \item Private/Shared: \char`~17:30 min
\end{itemize}

%\section{Example Appendix}
%\label{sec:appendix}

%This is an appendix.

\section{Additional Results}
\label{sec:sup-plots}

\begin{figure*}[h!]
    \centering
\begin{tabular}{cc}
    \centering
  \includegraphics[width=0.45\textwidth]{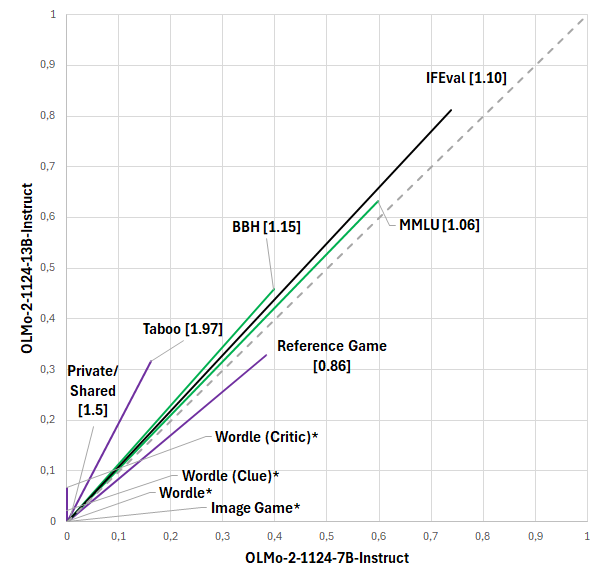}  &
   
  \includegraphics[width=0.45\textwidth]{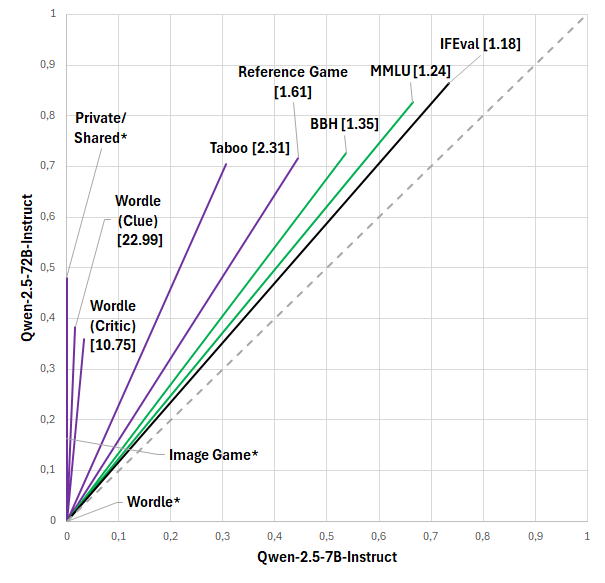}    
\end{tabular}\caption{Comparing datasets in their power to discriminate models across size.}\label{fig:additional_disc}
  \end{figure*}

\begin{figure*}[!p]
    \centering

\begin{tabular}{cc}
    \centering
  \includegraphics[width=0.45\textwidth]{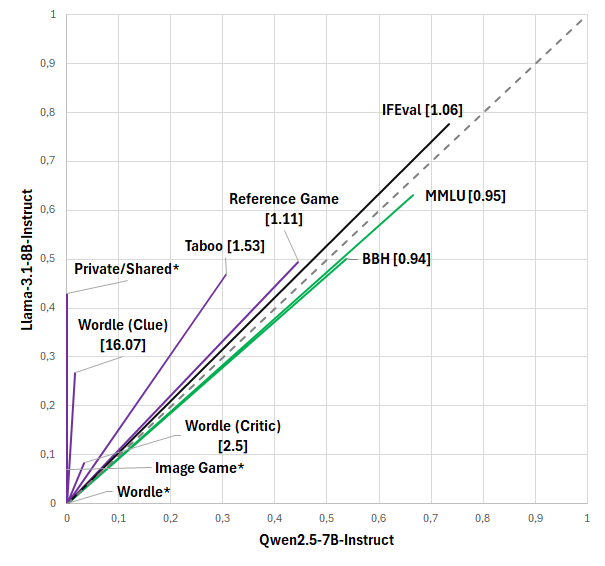}  & 
  \includegraphics[width=0.45\textwidth]{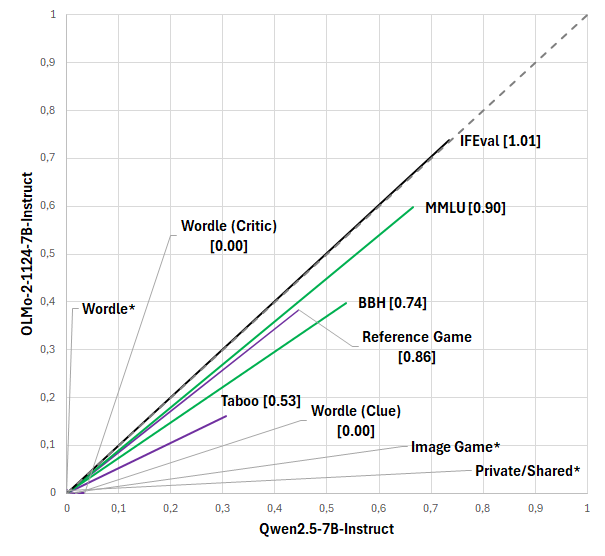} \\  
  \includegraphics[width=0.45\textwidth]{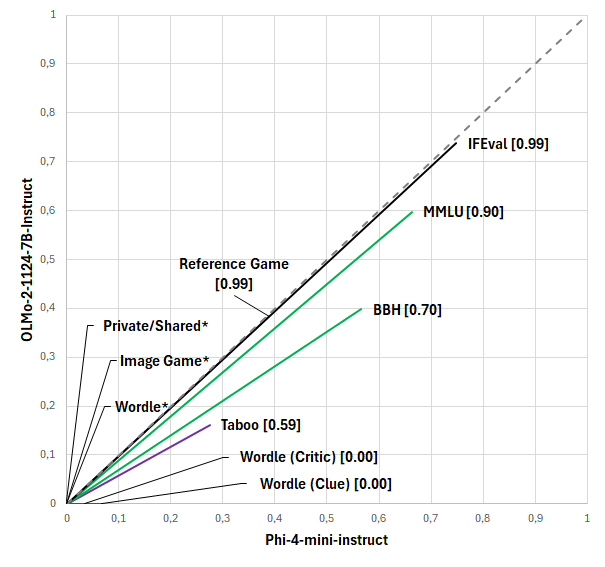} &  
  \includegraphics[width=0.45\textwidth]{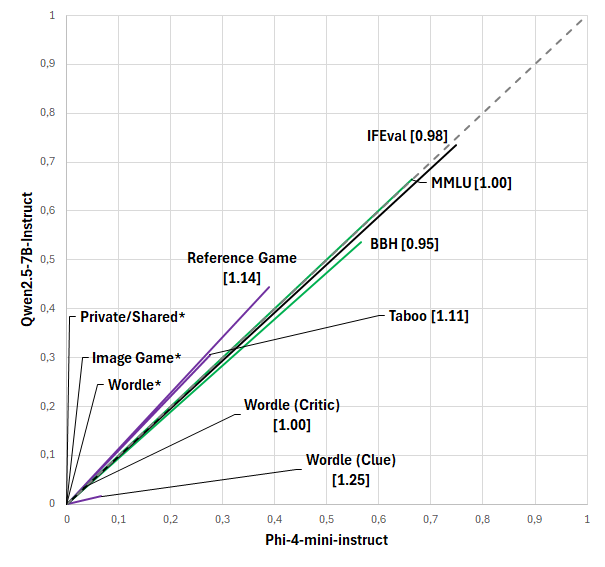}
\end{tabular}\caption{Comparing datasets in their power to discriminate models with similar size across families (Part 1).}\label{fig:additional_disc_same_size}
  \end{figure*}

\begin{figure*}[h!]
    \centering

\begin{tabular}{cc}
    \centering
  \includegraphics[width=0.45\textwidth]{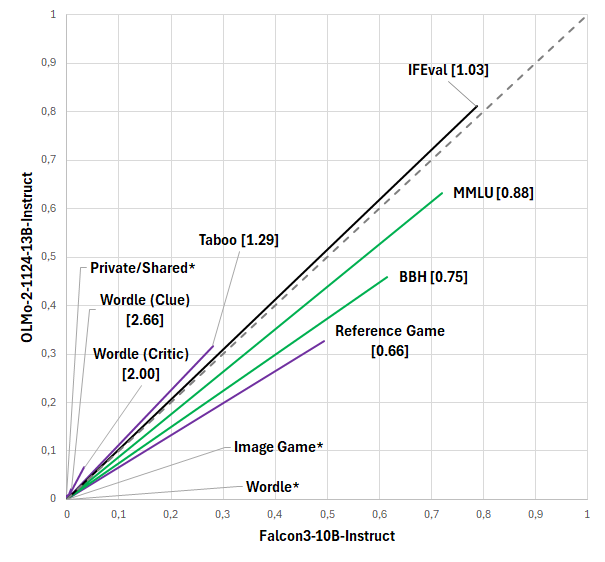} &
  \includegraphics[width=0.45\textwidth]{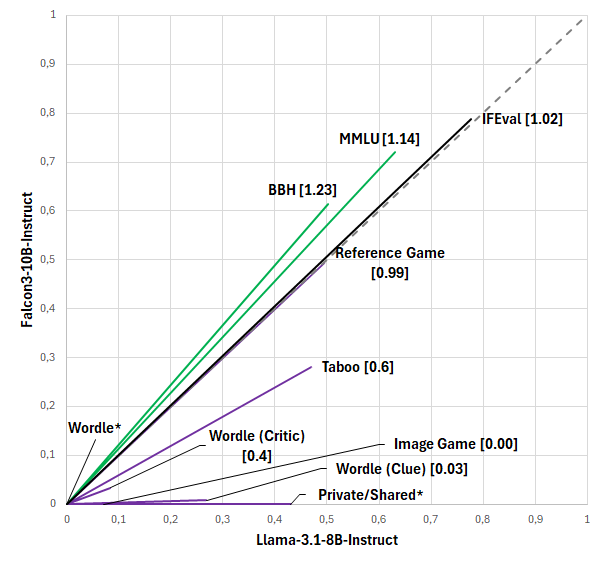}  \\
  \includegraphics[width=0.45\textwidth]{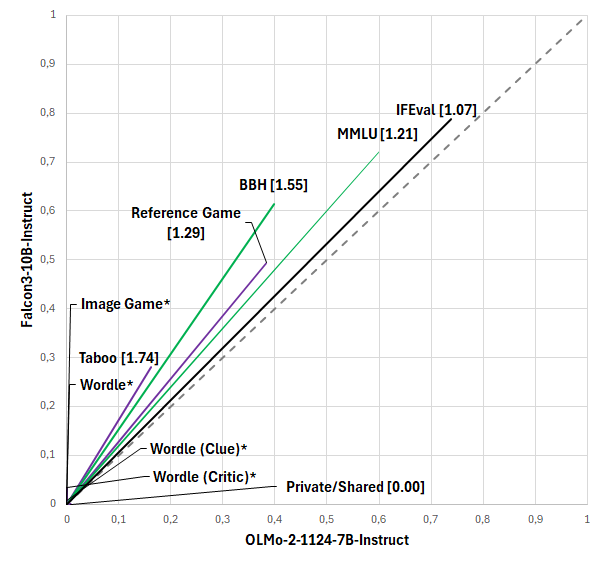} &  
  \includegraphics[width=0.45\textwidth]{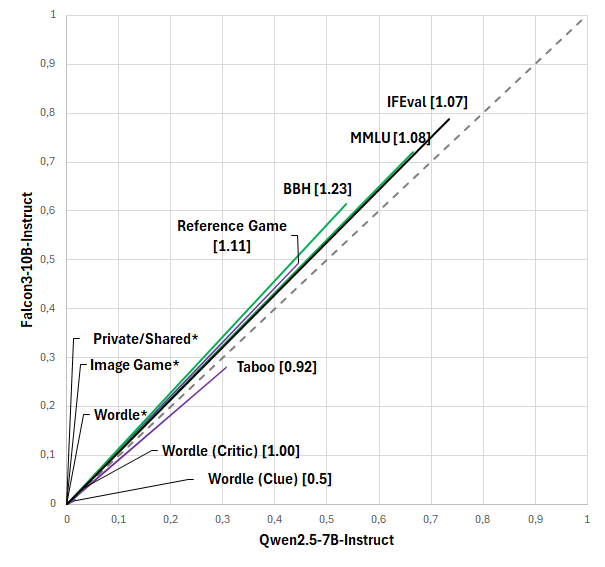} 
\end{tabular}\caption{Comparing datasets in their power to discriminate models with similar size across families (Part 2).}\label{fig:additional_disc_same_size_2}
  \end{figure*}

\begin{figure}[!ht]
    \centering
  \includegraphics[width=0.47\textwidth]{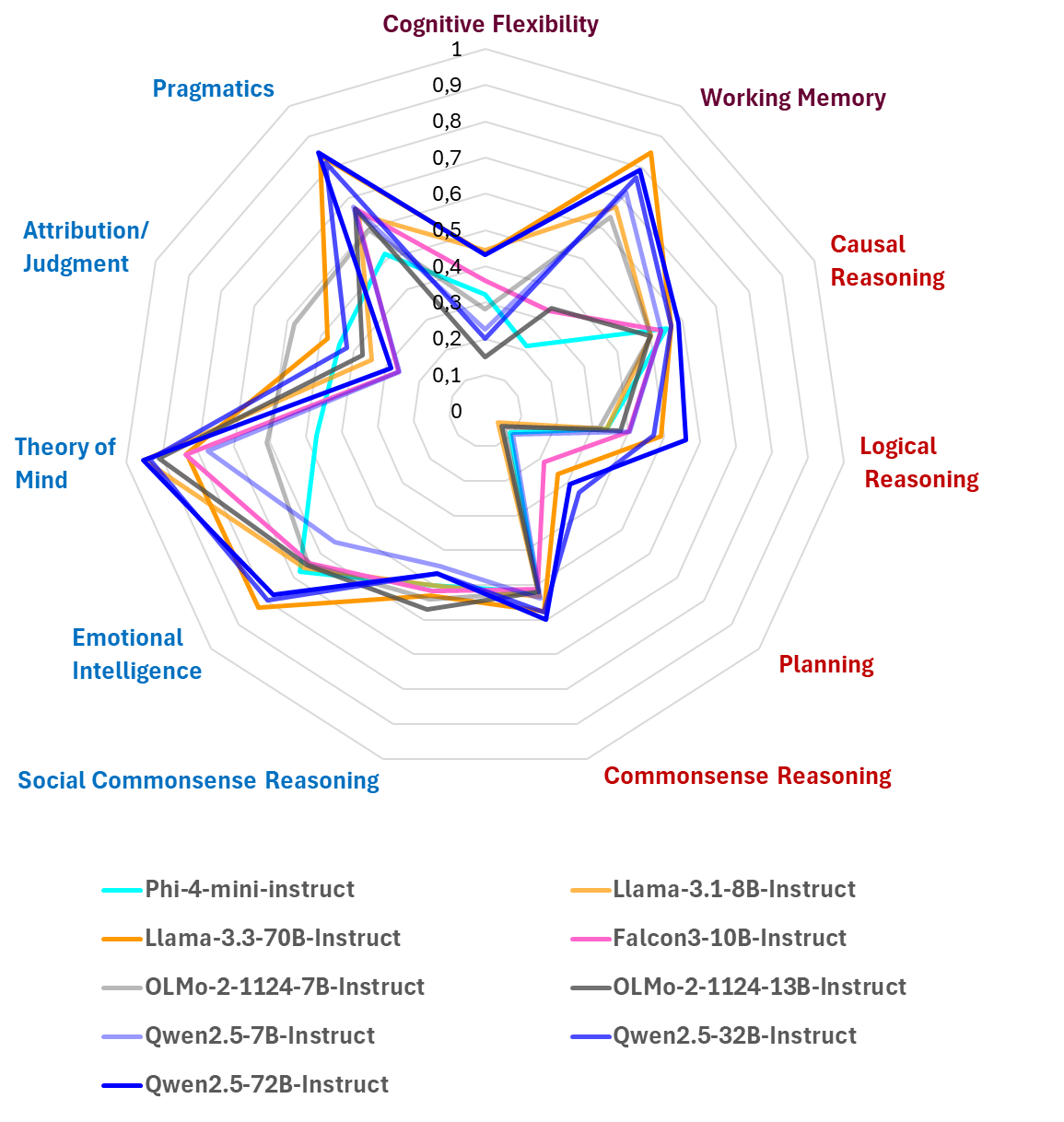}    \caption{Cognitive Abilities Spectrum of LLMs}  \label{fig:cogabilities_all}
  \end{figure}

\begin{figure}[!ht]
    \centering
    \begin{minipage}{\textwidth}
        \centering
        \begin{tabular}{cc}
            \includegraphics[width=0.48\textwidth]{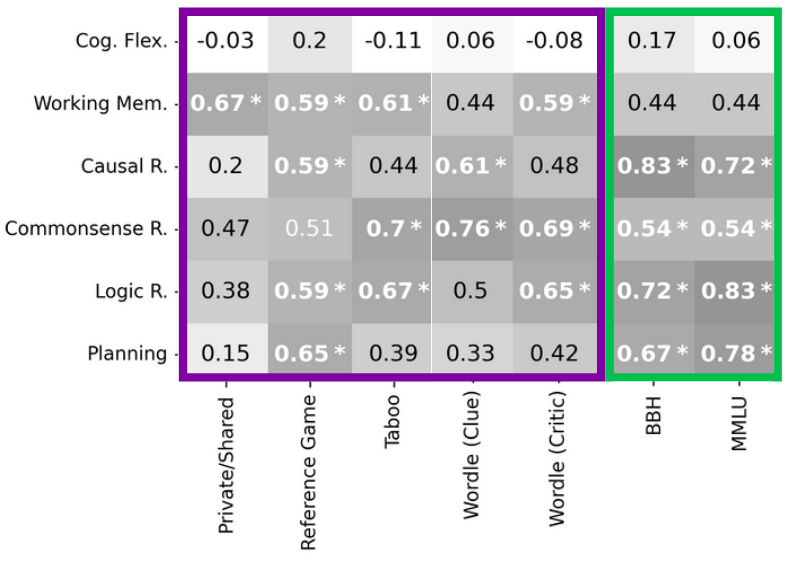} &
            \includegraphics[width=0.48\textwidth]{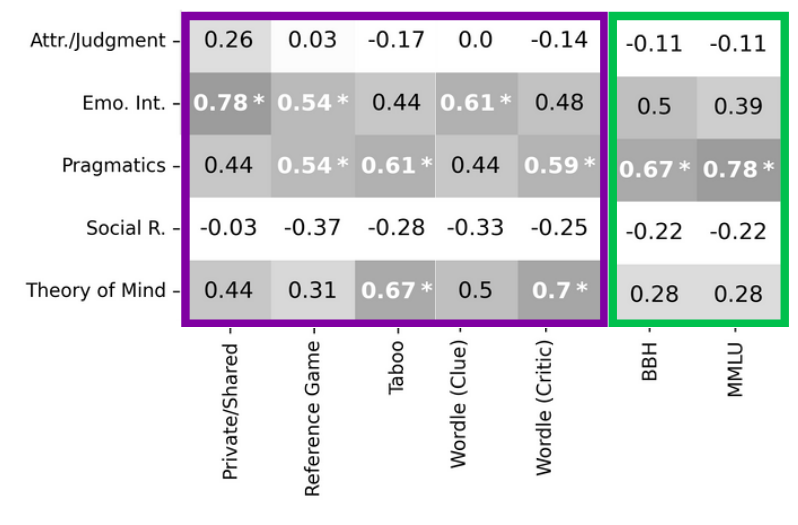}
        \end{tabular}
        \captionof{figure}{Correlation of social/emotional cognitive abilities with \textcolor{interactive_games}{Interactive Games} and \textcolor{massive_benchmarks}{Large QA benchmarks}. \\
        The correlation matrix does not include results on Wordle and Image Game as model performances were too low.}
        \label{fig:cogabilities_complete}
    \end{minipage}
\end{figure}

\begin{figure}[h!]
    \centering
    \begin{tabular}{cc}
 \includegraphics[width=0.5\textwidth]{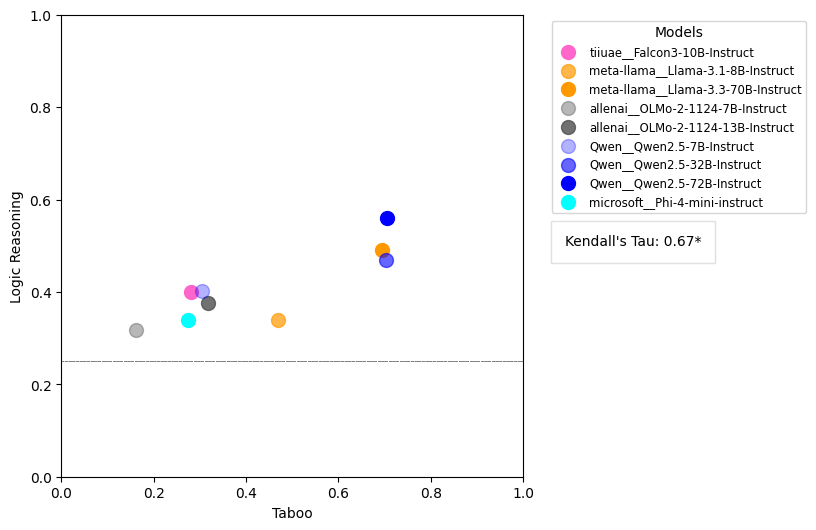} &
 \includegraphics[width=0.5\textwidth]{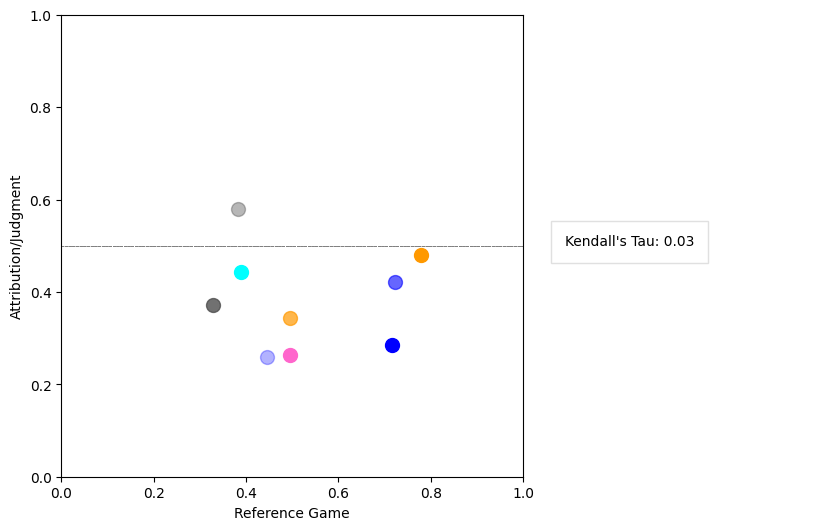}
\end{tabular}
   \caption{High (left) and low (right) correlation }  \label{fig:scatterplots}
\end{figure}

%\subsection{Static vs.}
Table~\ref{tab:tab-all} includes the scores achieved by every model on each benchmark.
Figure~\ref{fig:additional_disc}, Figure~\ref{fig:additional_disc_same_size} and Figure~\ref{fig:additional_disc_same_size_2} show the supplementary plots for the results in Section~\ref{sec:static-interactive} (comparing models of different size but same family and models of similar size but coming from different families, respectively). Moreover, we provide supplementary plots for the results in Section~\ref{sec:eval_ca}. Figure~\ref{fig:cogabilities_all} presents a direct comparison of models based on our selected cognitive tests.   Figure~\ref{fig:cogabilities_complete} reports an extended version of Figure~\ref{fig:cogabilities}. Finally, Figure~\ref{fig:scatterplots} reports two scatter plots exemplifying, respectively, situations of high and low correlation between two benchmarks (a game-based one and a cognitive ability one).

%Among higher-order executive functions, all models heavily struggle with planning and causal reasoning, though larger models perform better than smaller ones.
%While larger models exhibit some degree of working memory, they lack cognitive flexibility to the same extent as smaller models. In the domain of social and emotional abilities, the weakest area for all models is attribute/judgement (or applied theory of mind). 

\begin{table*}[htbp]
\centering
\small
\resizebox{\textwidth}{!}{
\begin{tabular}{lcccccccccc}
\toprule
\textbf{Metric} & \textbf{OLMo-2-7B} & \textbf{OLMo-2-13B} & \textbf{Llama-3.1-8B} & \textbf{Llama-3.3-70B} & \textbf{Qwen2.5-7B} & \textbf{Qwen2.5-32B} & \textbf{Qwen2.5-72B} & \textbf{Falcon3-10B} & \textbf{Phi-4-mini} \\
\midrule
Cognitive Flexibility & 0.2821 & 0.1500 & 0.4438 & 0.4339 & 0.2265 & 0.2008 & 0.4318 & 0.3615 & 0.3217 \\
\midrule
Working Memory & 0.6383 & 0.3383 & 0.6692 & 0.8475 & 0.7175 & 0.7683 & 0.7900 & 0.3292 & 0.2125 \\
\midrule
Causal Reasoning & 0.5034 & 0.4997 & 0.5035 & 0.5635 & 0.5329 & 0.5654 & 0.5856 & 0.5343 & 0.5489 \\
\midrule
Logical Reasoning & 0.3181 & 0.3766 & 0.3391 & 0.4905 & 0.4027 & 0.4701 & 0.5592 & 0.4001 & 0.3384 \\
\midrule
Planning & 0.0708 & 0.0613 & 0.0482 & 0.2666 & 0.0993 & 0.3428 & 0.3079 & 0.2147 & 0.0894 \\
\midrule
Commonsense Reasoning & 0.5193 & 0.5193 & 0.5328 & 0.5754 & 0.5375 & 0.5793 & 0.5983 & 0.5107 & 0.5201 \\
\midrule
Social Reasoning & 0.5404 & 0.5701 & 0.5000 & 0.5302 & 0.4458 & 0.4667 & 0.4672 & 0.5174 & 0.5015 \\
\midrule
Emotional Intelligence & 0.6456 & 0.6484 & 0.6614 & 0.8290 & 0.5504 & 0.7949 & 0.7752 & 0.6405 & 0.6776 \\
\midrule
Theory of Mind & 0.6085 & 0.9093 & 0.9503 & 0.8326 & 0.7742 & 0.9364 & 0.9538 & 0.8361 & 0.4708 \\
\midrule
Attribution/Judgment & 0.5803 & 0.3710 & 0.3444 & 0.4795 & 0.2602 & 0.4207 & 0.2847 & 0.2642 & 0.4429 \\
\midrule
Pragmatics & 0.5951 & 0.6634 & 0.6463 & 0.8402 & 0.6707 & 0.8146 & 0.8500 & 0.6646 & 0.5146 \\
\midrule
Taboo & 0.1611 & 0.3167 & 0.4694 & 0.6944 & 0.3056 & 0.7028 & 0.7056 & 0.2806 & 0.2750 \\
\midrule
Reference Game & 0.3833 & 0.3278 & 0.4944 & 0.7778 & 0.4444 & 0.7222 & 0.7167 & 0.4944 & 0.3889 \\
\midrule
Private/Shared & 0.0067 & 0.0100 & 0.4292 & 0.8985 & 0 & 0.4851 & 0.4799 & 0 & 0 \\
\midrule
Image Game & 0 & 0 & 0.0708 & 0.4710 & 0 & 0 & 0.1625 & 0 & 0 \\
\midrule
Wordle & 0 & 0 & 0 & 0.0300 & 0 & 0.0500 & 0 & 0 & 0 \\
\midrule
Wordle (Clue) & 0 & 0.0222 & 0.2678 & 0.2994 & 0.0167 & 0.3000 & 0.3833 & 0.0083 & 0.0667 \\
\midrule
Wordle (Critic) & 0 & 0.0667 & 0.0833 & 0.2500 & 0.0333 & 0.3167 & 0.3583 & 0.0333 & 0.0333 \\
\midrule
MMLU & 0.5973 & 0.6332 & 0.6305 & 0.7761 & 0.6651 & 0.7408 & 0.8264 & 0.7198 & 0.6627 \\
\midrule
BBH & 0.3980 & 0.4588 & 0.5013 & 0.6915 & 0.5369 & 0.6874 & 0.7256 & 0.6148 & 0.5662 \\
\midrule
IFEval & 0.7378 & 0.8114 & 0.7758 & 0.9098 & 0.7344 & 0.8322 & 0.8644 & 0.7878 & 0.7485 \\
\bottomrule
\end{tabular}
}
\caption{Scores obtained by models on each benchmark. The "Instruct" has been omitted from the column names for stylistic purposes.}
\label{tab:tab-all}
\end{table*}
\clearpage

\begin{figure}[b]
{ \footnotesize  \setcounter{utterance}{1}
\setlength{\tabcolsep}{0pt}
\begin{supertabular}{c@{$\;$}|p{.15\linewidth}@{}p{.15\linewidth}p{.15\linewidth}p{.15\linewidth}p{.15\linewidth}p{.15\linewidth}}

    \# & $\;$A & \multicolumn{4}{c}{Game Master} & $\;\:$B\\
    \hline 

    \theutterance \stepcounter{utterance}  

    & & \multicolumn{4}{p{0.6\linewidth}}{\cellcolor[rgb]{0.9,0.9,0.9}{%
	\makecell[{{p{\linewidth}}}]{% 
	  \tt {\tiny [A$\langle$GM]}  
	[initial prompt (describer)]\\ \tt \\ \tt This is the target word that you need to describe and that the other player needs to guess:\\ \tt \\ \tt clear\\ \tt \\ \tt Related words are:\\ \tt \\ \tt - crystal\\ \tt - obvious\\ \tt - confident\\ \tt \\ \tt Important: You are under time pressure, give short descriptions that are to the point! 
	  } 
	   } 
	   } 
	 & \\ \\

    \theutterance \stepcounter{utterance}  

    & \multicolumn{4}{p{0.6\linewidth}}{\cellcolor[rgb]{0.8,1,0.9}{%
	\makecell[{{p{\linewidth}}}]{% 
	  \tt {\tiny [A$\rangle$GM]}  
	 CLUE: Transparent and easy to see through. 
	  } 
	   } 
	   } 
	 & & \\

    \theutterance \stepcounter{utterance}  

    & & \multicolumn{4}{p{0.6\linewidth}}{\cellcolor[rgb]{0.9,0.9,0.9}{%
	\makecell[{{p{\linewidth}}}]{% 
	  \tt {\tiny [GM$\rangle$B]}  
	 [initial prompt (guesser)]\\ \tt \\ \tt \\ \tt CLUE: Transparent and easy to see through. 
	  } 
	   } 
	   } 
	 & \\

    \theutterance \stepcounter{utterance}  

    & & & \multicolumn{4}{p{0.6\linewidth}}{\cellcolor[rgb]{1,0.85,0.72}{%
	\makecell[{{p{\linewidth}}}]{% 
	  \tt {\tiny [GM$\langle$B]}  
	 GUESS: Glass 
	  } 
	   } 
	   } 
	  \\

    \theutterance \stepcounter{utterance}  

    & & \multicolumn{4}{p{0.6\linewidth}}{\cellcolor[rgb]{0.9,0.9,0.9}{%
	\makecell[{{p{\linewidth}}}]{% 
	  \tt {\tiny [A$\langle$GM]}  
	 GUESS: Glass 
	  } 
	   } 
	   } 
	 & \\

    \theutterance \stepcounter{utterance}  

    & \multicolumn{4}{p{0.6\linewidth}}{\cellcolor[rgb]{0.8,1,0.9}{%
	\makecell[{{p{\linewidth}}}]{% 
	  \tt {\tiny [A$\rangle$GM]}  
	 CLUE: Not a material, but a state of something, like air or water. 
	  } 
	   } 
	   } 
	 & & \\

    \theutterance \stepcounter{utterance}  

    & & \multicolumn{4}{p{0.6\linewidth}}{\cellcolor[rgb]{0.9,0.9,0.9}{%
	\makecell[{{p{\linewidth}}}]{% 
	  \tt {\tiny [GM$\rangle$B]}  
	 CLUE: Not a material, but a state of something, like air or water. 
	  } 
	   } 
	   } 
	 & \\

    \theutterance \stepcounter{utterance}  

    & & & \multicolumn{4}{p{0.6\linewidth}}{\cellcolor[rgb]{1,0.85,0.72}{%
	\makecell[{{p{\linewidth}}}]{% 
	  \tt {\tiny [GM$\langle$B]}  
	 GUESS: Clear 
	  } 
	   } 
	   } 
	  \\ 

\end{supertabular}
}
\caption{An example of an episode of Taboo played by Llama-3.3-70B-Instruct on both roles. The two players' messages are in green and orange, while the grey-colored text comes from the environment itself, represented by the Game Master.}
\label{taboo_transcript}
\end{figure}

\begin{figure}[b]
{ \footnotesize  \setcounter{utterance}{1}
\setlength{\tabcolsep}{0pt}
\begin{supertabular}{c@{$\;$}|p{.15\linewidth}@{}p{.15\linewidth}p{.15\linewidth}p{.15\linewidth}p{.15\linewidth}p{.15\linewidth}}

    \# & $\;$A & \multicolumn{4}{c}{Game Master} & $\;\:$B\\
    \hline 

    \theutterance \stepcounter{utterance}  
    & & \multicolumn{4}{p{0.6\linewidth}}{\cellcolor[rgb]{0.9,0.9,0.9}{%
	\makecell[{{p{\linewidth}}}]{% 
	  \tt {\tiny [A$\langle$GM]}  
	 [instruction prompt..]
	  } 
	   } 
	   } 
	 & \\

    \theutterance \stepcounter{utterance}  
    & \multicolumn{4}{p{0.6\linewidth}}{\cellcolor[rgb]{0.8,1,0.9}{%
	\makecell[{{p{\linewidth}}}]{% 
	  \tt {\tiny [A$\rangle$GM]}  
	 guess: house\\ \tt explanation: this is a common five-letter English word, and I am starting my guess with this word because it contains a mix of vowels and consonants that are frequently used in many words. 
	  } 
	   } 
	   } 
	 & & \\

    \theutterance \stepcounter{utterance}  
    & & \multicolumn{4}{p{0.6\linewidth}}{\cellcolor[rgb]{0.9,0.9,0.9}{%
	\makecell[{{p{\linewidth}}}]{% 
	  \tt {\tiny [A$\langle$GM]}  
	 guess\_feedback: h\textless yellow\textgreater\ o\textless red\textgreater\ u\textless red\textgreater\ s\textless red\textgreater\ e\textless green\textgreater\\ \tt \\ \tt Provide your response only in this format.\\ \tt guess: word\\ \tt explanation: details 
	  } 
	   } 
	   } 
	 & \\

    \theutterance \stepcounter{utterance}  
    & & & \multicolumn{2}{p{0.3\linewidth}}{\cellcolor[rgb]{0.95,0.95,0.95}{%
	\makecell[{{p{\linewidth}}}]{% 
	  \tt {\tiny [GM$|$GM]}  
	 [...]
	  } 
	   } 
	   } 
	 & & \\ 

    \theutterance \stepcounter{utterance}  
    & \multicolumn{4}{p{0.6\linewidth}}{\cellcolor[rgb]{0.8,1,0.9}{%
	\makecell[{{p{\linewidth}}}]{% 
	  \tt {\tiny [A$\rangle$GM]}  
	 guess: whine\\ \tt explanation: I'm replacing the "t" with a "w" to form a new word, as the letters "h", "i", "n", and "e" were all marked green, indicating they are in the correct positions, and "whine" is a valid English word that fits this pattern. 
	  } 
	   } 
	   } 
	 & & \\ 

    \theutterance \stepcounter{utterance}  
    & & & \multicolumn{2}{p{0.3\linewidth}}{\cellcolor[rgb]{0.95,0.95,0.95}{%
	\makecell[{{p{\linewidth}}}]{% 
	  \tt {\tiny [GM$|$GM]}  
	 game\_result = WIN 
	  } 
	   } 
	   } 
	 & & \\ 

\end{supertabular}
}

\caption{An example of an episode of Wordle played by Llama-3.3-70B-Instruct. The player's messages are in green, while the grey-colored text comes from the environment itself, represented by the Game Master.}
\label{wordle_transcript}
\end{figure}

\begin{figure}[p!]
{ \footnotesize  \setcounter{utterance}{1}
\setlength{\tabcolsep}{0pt}
\begin{supertabular}{c@{$\;$}|p{.15\linewidth}@{}p{.15\linewidth}p{.15\linewidth}p{.15\linewidth}p{.15\linewidth}p{.15\linewidth}}

    \# & $\;$A & \multicolumn{4}{c}{Game Master} & $\;\:$B\\
    \hline 

    \theutterance \stepcounter{utterance}  
    & & \multicolumn{4}{p{0.6\linewidth}}{\cellcolor[rgb]{0.9,0.9,0.9}{%
	\makecell[{{p{\linewidth}}}]{% 
	  \tt {\tiny [A$\langle$GM]}  
      [instruction prompt..]\\ \tt clue: complain
	  } 
	   } 
	   } 
	 & \\

    \theutterance \stepcounter{utterance}  
    & \multicolumn{4}{p{0.6\linewidth}}{\cellcolor[rgb]{0.8,1,0.9}{%
	\makecell[{{p{\linewidth}}}]{% 
	  \tt {\tiny [A$\rangle$GM]}  
	 guess: moans\\ \tt explanation: The clue "complain" makes me think of words related to expressing dissatisfaction or discomfort, and "moans" is a word that fits this description and has 5 letters, making it a plausible initial guess. 
	  } 
	   } 
	   } 
	 & & \\

    \theutterance \stepcounter{utterance}  
    & & \multicolumn{4}{p{0.6\linewidth}}{\cellcolor[rgb]{0.9,0.9,0.9}{%
	\makecell[{{p{\linewidth}}}]{% 
	  \tt {\tiny [A$\langle$GM]}  
	 guess\_feedback: m\textless red\textgreater\ o\textless red\textgreater\ a\textless red\textgreater\ n\textless green\textgreater\ s\textless red\textgreater\\ \tt \\ \tt Provide your response only in this format.\\ \tt guess: word\\ \tt explanation: details 
	  } 
	   } 
	   } 
	 & \\ 
 
    \theutterance \stepcounter{utterance}  
    & & & \multicolumn{2}{p{0.3\linewidth}}{\cellcolor[rgb]{0.95,0.95,0.95}{%
	\makecell[{{p{\linewidth}}}]{% 
	  \tt {\tiny [GM$|$GM]}  
	 [..]
	  } 
	   } 
	   } 
	 & & \\

    \theutterance \stepcounter{utterance}  
    & \multicolumn{4}{p{0.6\linewidth}}{\cellcolor[rgb]{0.8,1,0.9}{%
	\makecell[{{p{\linewidth}}}]{% 
	  \tt {\tiny [A$\rangle$GM]}  
	 guess: whine\\ \tt explanation: The clue "complain" and the feedback that both "i" and "n" are in the correct positions suggest a word related to expressing dissatisfaction, and "whine" is a word that fits this description, incorporating the correctly positioned "i" and "n", and having 5 letters, making it a plausible next guess. 
	  } 
	   } 
	   } 
	 & & \\ 
 
    \theutterance \stepcounter{utterance}  
    & & & \multicolumn{2}{p{0.3\linewidth}}{\cellcolor[rgb]{0.95,0.95,0.95}{%
	\makecell[{{p{\linewidth}}}]{% 
	  \tt {\tiny [GM$|$GM]}  
	 game\_result = WIN 
	  } 
	   } 
	   } 
	 & & \\ 

\end{supertabular}
}
\caption{An example of an episode of Wordle (variant with Clue) played by Llama-3.3-70B-Instruct. The player's messages are in green, while the grey-colored text comes from the environment itself, represented by the Game Master.}
\label{wordleclue_transcripts}
\end{figure}

\begin{figure}[p]
{ \footnotesize  \setcounter{utterance}{1}
\setlength{\tabcolsep}{0pt}
\begin{supertabular}{c@{$\;$}|p{.15\linewidth}@{}p{.15\linewidth}p{.15\linewidth}p{.15\linewidth}p{.15\linewidth}p{.15\linewidth}}

    \# & $\;$A & \multicolumn{4}{c}{Game Master} & $\;\:$B\\
    \hline 

    \theutterance \stepcounter{utterance}  

    & & \multicolumn{4}{p{0.6\linewidth}}{\cellcolor[rgb]{0.9,0.9,0.9}{%
	\makecell[{{p{\linewidth}}}]{% 
	  \tt {\tiny [A$\langle$GM]}  
	 [instruction prompt..]\ \tt clue: complain\\ \tt  
	  } 
	   } 
	   } 
	 & \\

    \theutterance \stepcounter{utterance}  

    & \multicolumn{4}{p{0.6\linewidth}}{\cellcolor[rgb]{0.8,1,0.9}{%
	\makecell[{{p{\linewidth}}}]{% 
	  \tt {\tiny [A$\rangle$GM]}  
	 guess: moans\\ \tt explanation: The word "moans" is related to complaining, as people often moan when they are unhappy or dissatisfied with something, and it fits the 5-letter length. 
	  } 
	   } 
	   } 
	 & & \\

    \theutterance \stepcounter{utterance}  

    & & \multicolumn{4}{p{0.6\linewidth}}{\cellcolor[rgb]{0.9,0.9,0.9}{%
	\makecell[{{p{\linewidth}}}]{% 
	  \tt {\tiny [GM$\rangle$B]}  
	 [instruction prompt..]
	  } 
	   } 
	   } 
	 & \\

    \theutterance \stepcounter{utterance}  

    & & & \multicolumn{4}{p{0.6\linewidth}}{\cellcolor[rgb]{1,0.85,0.72}{%
	\makecell[{{p{\linewidth}}}]{% 
	  \tt {\tiny [GM$\langle$B]}  
	 agreement: yes\\ \tt explanation: the guess "moans" is a plausible match for the clue "complain" because moaning is a way of expressing dissatisfaction or unhappiness, which is a form of complaining, and it meets the 5-letter length requirement. 
	  } 
	   } 
	   } 
	  \\

    \theutterance \stepcounter{utterance}  

    & & & \multicolumn{2}{p{0.3\linewidth}}{\cellcolor[rgb]{0.95,0.95,0.95}{%
	\makecell[{{p{\linewidth}}}]{% 
	  \tt {\tiny [GM$|$GM]}  
	 [..]
	  } 
	   } 
	   } 
	 & & \\

    \theutterance \stepcounter{utterance}  

    & \multicolumn{4}{p{0.6\linewidth}}{\cellcolor[rgb]{0.8,1,0.9}{%
	\makecell[{{p{\linewidth}}}]{% 
	  \tt {\tiny [A$\rangle$GM]}  
	 guess: whine\\ \tt explanation: The word "whine" is a form of complaining, often through a high-pitched or nasal sound, and it contains the letter "n" which was indicated by the feedback to be in the word, making it a potential match for the clue "complain" with a 5-letter length. 
	  } 
	   } 
	   } 
	 & & \\

    \theutterance \stepcounter{utterance}  

    & & & \multicolumn{2}{p{0.3\linewidth}}{\cellcolor[rgb]{0.95,0.95,0.95}{%
	\makecell[{{p{\linewidth}}}]{% 
	  \tt {\tiny [GM$|$GM]}  
	 game\_result = WIN 
	  } 
	   } 
	   }
	 & & \\

\end{supertabular}
}
\caption{An example of an episode of Wordle (variant with Critic) played by Llama-3.3-70B-Instruct on both roles. The two players' messages are in green and orange, while the grey-colored text comes from the environment itself, represented by the Game Master.}
\label{wordle_critic_transcript}
\end{figure}

\begin{figure}[p]
{ \footnotesize  \setcounter{utterance}{1}
\setlength{\tabcolsep}{0pt}
\begin{supertabular}{c@{$\;$}|p{.15\linewidth}@{}p{.15\linewidth}p{.15\linewidth}p{.15\linewidth}p{.15\linewidth}p{.15\linewidth}}

    \# & $\;$A & \multicolumn{4}{c}{Game Master} & $\;\:$B\\
    \hline 

    \theutterance \stepcounter{utterance}  

    & & \multicolumn{4}{p{0.6\linewidth}}{\cellcolor[rgb]{0.9,0.9,0.9}{%
	\makecell[{{p{\linewidth}}}]{% 
	  \tt {\tiny [A$\langle$GM]}  
	 [instruction prompt (describer)..] Ok. Please do this for the following example, which is a 5 by 5 grid.\\ \tt \\ \tt V V V V V\\ \tt $\square$ $\square$ $\square$ V $\square$\\ \tt $\square$ $\square$ V $\square$ $\square$\\ \tt $\square$ V $\square$ $\square$ $\square$\\ \tt V V V V V\\ \tt \\ \tt What is your next command?\\ \tt  
	  } 
	   } 
	   } 
	 & \\

    \theutterance \stepcounter{utterance}  

    & \multicolumn{4}{p{0.6\linewidth}}{\cellcolor[rgb]{0.8,1,0.9}{%
	\makecell[{{p{\linewidth}}}]{% 
	  \tt {\tiny [A$\rangle$GM]}  
	 Command: Fill the first row with V and fill the last row with V. 
	  } 
	   } 
	   } 
	 & & \\

    \theutterance \stepcounter{utterance}  

    & & \multicolumn{4}{p{0.6\linewidth}}{\cellcolor[rgb]{0.9,0.9,0.9}{%
	\makecell[{{p{\linewidth}}}]{% 
	  \tt {\tiny [GM$\rangle$B]}  
	 [instruction prompt (drawer)..]\\ \tt Command: Fill the first row with V and fill the last row with V. 
	  } 
	   } 
	   } 
	 & \\

    \theutterance \stepcounter{utterance}  

    & & & \multicolumn{4}{p{0.6\linewidth}}{\cellcolor[rgb]{1,0.85,0.72}{%
	\makecell[{{p{\linewidth}}}]{% 
	  \tt {\tiny [GM$\langle$B]}  
	 V V V V V\\ \tt $\square$ $\square$ $\square$ $\square$ $\square$\\ \tt $\square$ $\square$ $\square$ $\square$ $\square$\\ \tt $\square$ $\square$ $\square$ $\square$ $\square$\\ \tt V V V V V 
	  } 
	   } 
	   } 
	  \\

    \theutterance \stepcounter{utterance}  

    & & \multicolumn{4}{p{0.6\linewidth}}{\cellcolor[rgb]{0.9,0.9,0.9}{%
	\makecell[{{p{\linewidth}}}]{% 
	  \tt {\tiny [A$\langle$GM]}  
	 What is your next command? 
	  } 
	   } 
	   } 
	 & \\

    \theutterance \stepcounter{utterance}  

    & \multicolumn{4}{p{0.6\linewidth}}{\cellcolor[rgb]{0.8,1,0.9}{%
	\makecell[{{p{\linewidth}}}]{% 
	  \tt {\tiny [A$\rangle$GM]}  
	 Command: Put a V in the second row, fifth column, and put a V in the third row, third column, and put a V in the fourth row, second column. 
	  } 
	   } 
	   } 
	 & & \\

    \theutterance \stepcounter{utterance}  

    & & \multicolumn{4}{p{0.6\linewidth}}{\cellcolor[rgb]{0.9,0.9,0.9}{%
	\makecell[{{p{\linewidth}}}]{% 
	  \tt {\tiny [GM$\rangle$B]}  
	 Command: Put a V in the second row, fifth column, and put a V in the third row, third column, and put a V in the fourth row, second column. 
	  } 
	   } 
	   } 
	 & \\

    \theutterance \stepcounter{utterance}  

    & & & \multicolumn{4}{p{0.6\linewidth}}{\cellcolor[rgb]{1,0.85,0.72}{%
	\makecell[{{p{\linewidth}}}]{% 
	  \tt {\tiny [GM$\langle$B]}  
	 V V V V V\\ \tt $\square$ $\square$ $\square$ $\square$ V\\ \tt $\square$ $\square$ V $\square$ $\square$\\ \tt $\square$ V $\square$ $\square$ $\square$\\ \tt V V V V V 
	  } 
	   } 
	   } 
	  \\

    \theutterance \stepcounter{utterance}  

    & & \multicolumn{4}{p{0.6\linewidth}}{\cellcolor[rgb]{0.9,0.9,0.9}{%
	\makecell[{{p{\linewidth}}}]{% 
	  \tt {\tiny [A$\langle$GM]}  
	 What is your next command? 
	  } 
	   } 
	   } 
	 & \\

    \theutterance \stepcounter{utterance}  

    & \multicolumn{4}{p{0.6\linewidth}}{\cellcolor[rgb]{0.8,1,0.9}{%
	\makecell[{{p{\linewidth}}}]{% 
	  \tt {\tiny [A$\rangle$GM]}  
	 Command: DONE 
	  } 
	   } 
	   } 
	 & & \\

\end{supertabular}
}
\caption{An example of an episode of Image Game played by Llama-3.3-70B-Instruct on both roles. The two players' messages are in green and orange, while the grey-colored text comes from the environment itself, represented by the Game Master.}
\label{imagegame_transcripts}
\end{figure}

\begin{figure}[p]
  \centering
  {\footnotesize
    \setcounter{utterance}{1}
    \setlength{\tabcolsep}{0pt}
% Restrict content to one column using minipage
\noindent
\begin{minipage}{\columnwidth}
{ \scriptsize  \setcounter{utterance}{1}
\setlength{\tabcolsep}{1pt}
\begin{supertabular}{c@{$\;$}|p{0.12\linewidth}@{}p{0.12\linewidth}p{0.12\linewidth}p{0.12\linewidth}p{0.12\linewidth}p{0.12\linewidth}}

    \# & $\;$A & \multicolumn{4}{c}{Game Master} & $\;\:$B\\
    \hline 

    \theutterance \stepcounter{utterance}  
    & & \multicolumn{4}{p{0.6\linewidth}}{\cellcolor[rgb]{0.9,0.9,0.9}{%
	\makecell[{{p{\linewidth}}}]{% 
	  \tt {\tiny [A$\langle$GM]}  
	 [instruction prompt (describer)..]\\ \tt \\ \tt Target grid:\\ \tt \\ \tt X X $\square$ X X\\ \tt X X $\square$ X X\\ \tt $\square$ $\square$ $\square$ $\square$ $\square$\\ \tt X X $\square$ X X\\ \tt X X $\square$ X X\\ \tt \\ \tt Distractor grid 1:\\ \tt \\ \tt X X $\square$ X X\\ \tt X X $\square$ X X\\ \tt X X X X X\\ \tt X X $\square$ X X\\ \tt X X $\square$ X X\\ \tt \\ \tt Distractor grid 2:\\ \tt \\ \tt X X X X X\\ \tt X X X X X\\ \tt $\square$ $\square$ X $\square$ $\square$\\ \tt X X X X X\\ \tt X X X X X\\ \tt \\ \tt Instruction: Describe the target grid.\\ \tt Generate the referring expression starting with the tag "Expression: " for the given target grid. Omit any other text. 
	  } 
	   } 
	   } 
	 & \\ 

    \theutterance \stepcounter{utterance}  
    & \multicolumn{4}{p{0.6\linewidth}}{\cellcolor[rgb]{0.8,1,0.9}{%
	\makecell[{{p{\linewidth}}}]{% 
	  \tt {\tiny [A$\rangle$GM]}  
	 Expression: The grid with an empty row in the middle. 
	  } 
	   } 
	   } 
	 & & \\ 
    \theutterance \stepcounter{utterance}  
    & & \multicolumn{4}{p{0.6\linewidth}}{\cellcolor[rgb]{0.9,0.9,0.9}{%
	\makecell[{{p{\linewidth}}}]{% 
	  \tt {\tiny [GM$\rangle$B]}  
	 [instruction prompt (guesser)..]\\ \tt \\ \tt First:\\ \tt \\ \tt X X $\square$ X X\\ \tt X X $\square$ X X\\ \tt X X X X X\\ \tt X X $\square$ X X\\ \tt X X $\square$ X X\\ \tt \\ \tt Second:\\ \tt \\ \tt X X $\square$ X X\\ \tt X X $\square$ X X\\ \tt $\square$ $\square$ $\square$ $\square$ $\square$\\ \tt X X $\square$ X X\\ \tt X X $\square$ X X\\ \tt \\ \tt Third:\\ \tt \\ \tt X X X X X\\ \tt X X X X X\\ \tt $\square$ $\square$ X $\square$ $\square$\\ \tt X X X X X\\ \tt X X X X X\\ \tt \\ \tt Expression: The grid with an empty row in the middle.\\ \tt Question: Which grid does the expression refer to?\\ \tt Start with the tag "Answer: ", followed by your selection. Omit any other text.\\ \tt  
	  } 
	   } 
	   } 
	 & \\
    \theutterance \stepcounter{utterance}  
    & & & \multicolumn{4}{p{0.6\linewidth}}{\cellcolor[rgb]{1,0.85,0.72}{%
	\makecell[{{p{\linewidth}}}]{% 
	  \tt {\tiny [GM$\langle$B]}  
	 Answer: Second 
	  } 
	   } 
	   } 
	  \\ 

\end{supertabular}
}
\end{minipage}
}
  \caption{An example of an episode of Reference Game played by Llama-3.3-70B-Instruct on both roles. The two players' messages are in green and orange, while the grey-colored text comes from the environment itself, represented by the Game Master.}
  \label{reference_transcript}
\end{figure}

\begin{figure}[p]
{ \footnotesize  \setcounter{utterance}{1}
\setlength{\tabcolsep}{0pt}
\begin{supertabular}{c@{$\;$}|p{.15\linewidth}@{}p{.15\linewidth}p{.15\linewidth}p{.15\linewidth}p{.15\linewidth}p{.15\linewidth}}

    \# & $\;$A & \multicolumn{4}{c}{Game Master} & $\;\:$B\\
    \hline 

    \theutterance \stepcounter{utterance}  

    & & \multicolumn{4}{p{0.6\linewidth}}{\cellcolor[rgb]{0.9,0.9,0.9}{%
	\makecell[{{p{\linewidth}}}]{% 
	  \tt {\tiny [A$\langle$GM]}  
	 [instruction prompt..]
	  } 
	   } 
	   } 
	 & \\

    \theutterance \stepcounter{utterance}  

    & & \multicolumn{4}{p{0.6\linewidth}}{\cellcolor[rgb]{0.9,0.9,0.9}{%
	\makecell[{{p{\linewidth}}}]{% 
	  \tt {\tiny [GM$\rangle$B]}  
	 What is the next request? 
	  } 
	   } 
	   } 
	 & \\

    \theutterance \stepcounter{utterance}  

    & & & \multicolumn{4}{p{0.6\linewidth}}{\cellcolor[rgb]{1,0.85,0.72}{%
	\makecell[{{p{\linewidth}}}]{% 
	  \tt {\tiny [GM$\langle$B]}  
	 Where do you want to go? 
	  } 
	   } 
	   } 
	  \\

    \theutterance \stepcounter{utterance}  

    & & \multicolumn{4}{p{0.6\linewidth}}{\cellcolor[rgb]{0.9,0.9,0.9}{%
	\makecell[{{p{\linewidth}}}]{% 
	  \tt {\tiny [A$\langle$GM]}  
	 TRAVEL-AGENT: Where do you want to go? Answer only this question in one short sentence and then stop. 
	  } 
	   } 
	   } 
	 & \\

    \theutterance \stepcounter{utterance}  

    & \multicolumn{4}{p{0.6\linewidth}}{\cellcolor[rgb]{0.8,1,0.9}{%
	\makecell[{{p{\linewidth}}}]{% 
	  \tt {\tiny [A$\rangle$GM]}  
	 ANSWER: Dresden 
	  } 
	   } 
	   } 
	 & & \\

    \theutterance \stepcounter{utterance}  

    & & \multicolumn{4}{p{0.6\linewidth}}{\cellcolor[rgb]{0.9,0.9,0.9}{%
	\makecell[{{p{\linewidth}}}]{% 
	  \tt {\tiny [GM$\rangle$B]}  
	 Dresden 
	  } 
	   } 
	   } 
	 & \\

    \theutterance \stepcounter{utterance}  

    & & \multicolumn{4}{p{0.6\linewidth}}{\cellcolor[rgb]{0.9,0.9,0.9}{%
	\makecell[{{p{\linewidth}}}]{% 
	  \tt {\tiny [GM$\rangle$B]}  
	 What is the next request? 
	  } 
	   } 
	   } 
	 & \\

    \theutterance \stepcounter{utterance}  

    & & & \multicolumn{4}{p{0.6\linewidth}}{\cellcolor[rgb]{1,0.85,0.72}{%
	\makecell[{{p{\linewidth}}}]{% 
	  \tt {\tiny [GM$\langle$B]}  
	 How do you want to reach your destination? 
	  } 
	   } 
	   } 
	  \\

    \theutterance \stepcounter{utterance}  

    & & \multicolumn{4}{p{0.6\linewidth}}{\cellcolor[rgb]{0.9,0.9,0.9}{%
	\makecell[{{p{\linewidth}}}]{% 
	  \tt {\tiny [A$\langle$GM]}  
	 TRAVEL-AGENT: How do you want to reach your destination? [..] 
	  } 
	   } 
	   } 
	 & \\

    \theutterance \stepcounter{utterance}  

    & \multicolumn{4}{p{0.6\linewidth}}{\cellcolor[rgb]{0.8,1,0.9}{%
	\makecell[{{p{\linewidth}}}]{% 
	  \tt {\tiny [A$\rangle$GM]}  
	 ANSWER: By overnight bus. 
	  } 
	   } 
	   } 
	 & & \\

    \theutterance \stepcounter{utterance}  

    & & \multicolumn{4}{p{0.6\linewidth}}{\cellcolor[rgb]{0.9,0.9,0.9}{%
	\makecell[{{p{\linewidth}}}]{% 
	  \tt {\tiny [GM$\rangle$B]}  
	 By overnight bus. 
	  } 
	   } 
	   } 
	 & \\

    \theutterance \stepcounter{utterance}  

    & & \multicolumn{4}{p{0.6\linewidth}}{\cellcolor[rgb]{0.9,0.9,0.9}{%
	\makecell[{{p{\linewidth}}}]{% 
	  \tt {\tiny [GM$\rangle$B]}  
	 What is the next request? 
	  } 
	   } 
	   } 
	 & \\

    \theutterance \stepcounter{utterance}  

    & & & \multicolumn{4}{p{0.6\linewidth}}{\cellcolor[rgb]{1,0.85,0.72}{%
	\makecell[{{p{\linewidth}}}]{% 
	  \tt {\tiny [GM$\langle$B]}  
	 Where are you departing from? 
	  } 
	   } 
	   } 
	  \\

    \theutterance \stepcounter{utterance}  

    & & \multicolumn{4}{p{0.6\linewidth}}{\cellcolor[rgb]{0.9,0.9,0.9}{%
	\makecell[{{p{\linewidth}}}]{% 
	  \tt {\tiny [A$\langle$GM]}  
	 TRAVEL-AGENT: Where are you departing from? [..]
	  } 
	   } 
	   } 
	 & \\

    \theutterance \stepcounter{utterance}  

    & \multicolumn{4}{p{0.6\linewidth}}{\cellcolor[rgb]{0.8,1,0.9}{%
	\makecell[{{p{\linewidth}}}]{% 
	  \tt {\tiny [A$\rangle$GM]}  
	 ANSWER: Hamburg. 
	  } 
	   } 
	   } 
	 & & \\

    \theutterance \stepcounter{utterance}  

    & & \multicolumn{4}{p{0.6\linewidth}}{\cellcolor[rgb]{0.9,0.9,0.9}{%
	\makecell[{{p{\linewidth}}}]{% 
	  \tt {\tiny [GM$\rangle$B]}  
	 Hamburg. 
	  } 
	   } 
	   } 
	 & \\

    \theutterance \stepcounter{utterance}  

    & & \multicolumn{4}{p{0.6\linewidth}}{\cellcolor[rgb]{0.9,0.9,0.9}{%
	\makecell[{{p{\linewidth}}}]{% 
	  \tt {\tiny [GM$\rangle$B]}  
	 What is the next request? 
	  } 
	   } 
	   } 
	 & \\

    \theutterance \stepcounter{utterance}  

    & & & \multicolumn{4}{p{0.6\linewidth}}{\cellcolor[rgb]{1,0.85,0.72}{%
	\makecell[{{p{\linewidth}}}]{% 
	  \tt {\tiny [GM$\langle$B]}  
	 What kind of class do you wish? 
	  } 
	   } 
	   } 
	  \\

    \theutterance \stepcounter{utterance}  

    & & \multicolumn{4}{p{0.6\linewidth}}{\cellcolor[rgb]{0.9,0.9,0.9}{%
	\makecell[{{p{\linewidth}}}]{% 
	  \tt {\tiny [A$\langle$GM]}  
	 TRAVEL-AGENT: What kind of class do you wish? [..]
	  } 
	   } 
	   } 
	 & \\

    \theutterance \stepcounter{utterance}  

    & \multicolumn{4}{p{0.6\linewidth}}{\cellcolor[rgb]{0.8,1,0.9}{%
	\makecell[{{p{\linewidth}}}]{% 
	  \tt {\tiny [A$\rangle$GM]}  
	 ANSWER: The most comfortable. 
	  } 
	   } 
	   } 
	 & & \\

    \theutterance \stepcounter{utterance}  

    & & \multicolumn{4}{p{0.6\linewidth}}{\cellcolor[rgb]{0.9,0.9,0.9}{%
	\makecell[{{p{\linewidth}}}]{% 
	  \tt {\tiny [GM$\rangle$B]}  
	 The most comfortable. 
	  } 
	   } 
	   } 
	 & \\

    \theutterance \stepcounter{utterance}  

    & & \multicolumn{4}{p{0.6\linewidth}}{\cellcolor[rgb]{0.9,0.9,0.9}{%
	\makecell[{{p{\linewidth}}}]{% 
	  \tt {\tiny [GM$\rangle$B]}  
	 What is the next request? 
	  } 
	   } 
	   } 
	 & \\

    \theutterance \stepcounter{utterance}  

    & & & \multicolumn{4}{p{0.6\linewidth}}{\cellcolor[rgb]{1,0.85,0.72}{%
	\makecell[{{p{\linewidth}}}]{% 
	  \tt {\tiny [GM$\langle$B]}  
	 When does your trip begin? 
	  } 
	   } 
	   } 
	  \\

    \theutterance \stepcounter{utterance}  

    & & \multicolumn{4}{p{0.6\linewidth}}{\cellcolor[rgb]{0.9,0.9,0.9}{%
	\makecell[{{p{\linewidth}}}]{% 
	  \tt {\tiny [A$\langle$GM]}  
	 TRAVEL-AGENT: When does your trip begin? [..] 
	  } 
	   } 
	   } 
	 & \\

    \theutterance \stepcounter{utterance}  

    & \multicolumn{4}{p{0.6\linewidth}}{\cellcolor[rgb]{0.8,1,0.9}{%
	\makecell[{{p{\linewidth}}}]{% 
	  \tt {\tiny [A$\rangle$GM]}  
	 ANSWER: Third week of June. 
	  } 
	   } 
	   } 
	 & & \\

    \theutterance \stepcounter{utterance}  

    & & \multicolumn{4}{p{0.6\linewidth}}{\cellcolor[rgb]{0.9,0.9,0.9}{%
	\makecell[{{p{\linewidth}}}]{% 
	  \tt {\tiny [GM$\rangle$B]}  
	 Third week of June. 
	  } 
	   } 
	   } 
	 & \\

\end{supertabular}
}
\caption{An example of the main interaction of a Private/Shared episode played by Llama-3.3-70B-Instruct with a programmatic player. The two players' messages are in green and orange, while the grey-colored text comes from the environment itself, represented by the Game Master.}
\label{privateshared_transcripts}
\end{figure}

\begin{figure}[p]
{ \footnotesize  \setcounter{utterance}{1}
\setlength{\tabcolsep}{0pt}
\begin{supertabular}{c@{$\;$}|p{.15\linewidth}@{}p{.15\linewidth}p{.15\linewidth}p{.15\linewidth}p{.15\linewidth}p{.15\linewidth}}

    \# & $\;$A & \multicolumn{4}{c}{Game Master} & $\;\:$B\\
    \hline

    \theutterance \stepcounter{utterance}  

    & & & \multicolumn{2}{p{0.3\linewidth}}{\cellcolor[rgb]{0.95,0.95,0.95}{%
	\makecell[{{p{\linewidth}}}]{% 
	  \tt {\tiny [GM$|$GM]}  
	 Begin probing 
	  } 
	   } 
	   } 
	 & & \\

    \theutterance \stepcounter{utterance}  

    & & \multicolumn{4}{p{0.6\linewidth}}{\cellcolor[rgb]{0.9,0.9,0.9}{%
	\makecell[{{p{\linewidth}}}]{% 
	  \tt {\tiny [A$\langle$GM]}  
	 ME:  Do you think the travel agent knows where you depart from? Answer this question with yes or no and then stop. 
	  } 
	   } 
	   } 
	 & \\

    \theutterance \stepcounter{utterance}  

    & \multicolumn{4}{p{0.6\linewidth}}{\cellcolor[rgb]{0.8,1,0.9}{%
	\makecell[{{p{\linewidth}}}]{% 
	  \tt {\tiny [A$\rangle$GM]}  
	 ASIDE: Yes 
	  } 
	   } 
	   } 
	 & & \\

    \theutterance \stepcounter{utterance}  

    & & & \multicolumn{2}{p{0.3\linewidth}}{\cellcolor[rgb]{0.95,0.95,0.95}{%
	\makecell[{{p{\linewidth}}}]{% 
	  \tt {\tiny [GM$|$GM]}  
	 yes 
	  } 
	   } 
	   } 
	 & & \\

    \theutterance \stepcounter{utterance}  

    & & & \multicolumn{2}{p{0.3\linewidth}}{\cellcolor[rgb]{0.95,0.95,0.95}{%
	\makecell[{{p{\linewidth}}}]{% 
	  \tt {\tiny [GM$|$GM]}  
	 Answer for FROM valid after 1 tries. 
	  } 
	   } 
	   } 
	 & & \\

    \theutterance \stepcounter{utterance}  

    & & & \multicolumn{2}{p{0.3\linewidth}}{\cellcolor[rgb]{0.95,0.95,0.95}{%
	\makecell[{{p{\linewidth}}}]{% 
	  \tt {\tiny [GM$|$GM]}  
	 Answer is correct. 
	  } 
	   } 
	   } 
	 & & \\

    \theutterance \stepcounter{utterance}  

    & & \multicolumn{4}{p{0.6\linewidth}}{\cellcolor[rgb]{0.9,0.9,0.9}{%
	\makecell[{{p{\linewidth}}}]{% 
	  \tt {\tiny [A$\langle$GM]}  
	 ME:  Do you think the travel agent knows where you want to go to? [..] 
	  } 
	   } 
	   } 
	 & \\

    \theutterance \stepcounter{utterance}  

    & \multicolumn{4}{p{0.6\linewidth}}{\cellcolor[rgb]{0.8,1,0.9}{%
	\makecell[{{p{\linewidth}}}]{% 
	  \tt {\tiny [A$\rangle$GM]}  
	 ASIDE: Yes 
	  } 
	   } 
	   } 
	 & & \\

    \theutterance \stepcounter{utterance}  

    & & & \multicolumn{2}{p{0.3\linewidth}}{\cellcolor[rgb]{0.95,0.95,0.95}{%
	\makecell[{{p{\linewidth}}}]{% 
	  \tt {\tiny [GM$|$GM]}  
	 yes 
	  } 
	   } 
	   } 
	 & & \\

    \theutterance \stepcounter{utterance}  

    & & & \multicolumn{2}{p{0.3\linewidth}}{\cellcolor[rgb]{0.95,0.95,0.95}{%
	\makecell[{{p{\linewidth}}}]{% 
	  \tt {\tiny [GM$|$GM]}  
	 Answer for TO valid after 1 tries. 
	  } 
	   } 
	   } 
	 & & \\

    \theutterance \stepcounter{utterance}  

    & & & \multicolumn{2}{p{0.3\linewidth}}{\cellcolor[rgb]{0.95,0.95,0.95}{%
	\makecell[{{p{\linewidth}}}]{% 
	  \tt {\tiny [GM$|$GM]}  
	 Answer is correct. 
	  } 
	   } 
	   } 
	 & & \\

    \theutterance \stepcounter{utterance}  

    & & \multicolumn{4}{p{0.6\linewidth}}{\cellcolor[rgb]{0.9,0.9,0.9}{%
	\makecell[{{p{\linewidth}}}]{% 
	  \tt {\tiny [A$\langle$GM]}  
	 ME:  Is the travel agent aware of your means of transportation? [..] 
	  } 
	   } 
	   } 
	 & \\

    \theutterance \stepcounter{utterance}  

    & \multicolumn{4}{p{0.6\linewidth}}{\cellcolor[rgb]{0.8,1,0.9}{%
	\makecell[{{p{\linewidth}}}]{% 
	  \tt {\tiny [A$\rangle$GM]}  
	 ASIDE: Yes 
	  } 
	   } 
	   } 
	 & & \\

    \theutterance \stepcounter{utterance}  

    & & & \multicolumn{2}{p{0.3\linewidth}}{\cellcolor[rgb]{0.95,0.95,0.95}{%
	\makecell[{{p{\linewidth}}}]{% 
	  \tt {\tiny [GM$|$GM]}  
	 yes 
	  } 
	   } 
	   } 
	 & & \\

    \theutterance \stepcounter{utterance}  

    & & & \multicolumn{2}{p{0.3\linewidth}}{\cellcolor[rgb]{0.95,0.95,0.95}{%
	\makecell[{{p{\linewidth}}}]{% 
	  \tt {\tiny [GM$|$GM]}  
	 Answer for BY valid after 1 tries. 
	  } 
	   } 
	   } 
	 & & \\

    \theutterance \stepcounter{utterance}  

    & & & \multicolumn{2}{p{0.3\linewidth}}{\cellcolor[rgb]{0.95,0.95,0.95}{%
	\makecell[{{p{\linewidth}}}]{% 
	  \tt {\tiny [GM$|$GM]}  
	 Answer is correct. 
	  } 
	   } 
	   } 
	 & & \\

    \theutterance \stepcounter{utterance}  

    & & \multicolumn{4}{p{0.6\linewidth}}{\cellcolor[rgb]{0.9,0.9,0.9}{%
	\makecell[{{p{\linewidth}}}]{% 
	  \tt {\tiny [A$\langle$GM]}  
	 ME:  Have you informed the travel agent about your class preference? [..] 
	  } 
	   } 
	   } 
	 & \\

    \theutterance \stepcounter{utterance}  

    & \multicolumn{4}{p{0.6\linewidth}}{\cellcolor[rgb]{0.8,1,0.9}{%
	\makecell[{{p{\linewidth}}}]{% 
	  \tt {\tiny [A$\rangle$GM]}  
	 ASIDE: Yes 
	  } 
	   } 
	   } 
	 & & \\

    \theutterance \stepcounter{utterance}  

    & & & \multicolumn{2}{p{0.3\linewidth}}{\cellcolor[rgb]{0.95,0.95,0.95}{%
	\makecell[{{p{\linewidth}}}]{% 
	  \tt {\tiny [GM$|$GM]}  
	 yes 
	  } 
	   } 
	   } 
	 & & \\

    \theutterance \stepcounter{utterance}  

    & & & \multicolumn{2}{p{0.3\linewidth}}{\cellcolor[rgb]{0.95,0.95,0.95}{%
	\makecell[{{p{\linewidth}}}]{% 
	  \tt {\tiny [GM$|$GM]}  
	 Answer for CLASS valid after 1 tries. 
	  } 
	   } 
	   } 
	 & & \\

    \theutterance \stepcounter{utterance}  

    & & & \multicolumn{2}{p{0.3\linewidth}}{\cellcolor[rgb]{0.95,0.95,0.95}{%
	\makecell[{{p{\linewidth}}}]{% 
	  \tt {\tiny [GM$|$GM]}  
	 Answer is correct. 
	  } 
	   } 
	   } 
	 & & \\

    \theutterance \stepcounter{utterance}  

    & & \multicolumn{4}{p{0.6\linewidth}}{\cellcolor[rgb]{0.9,0.9,0.9}{%
	\makecell[{{p{\linewidth}}}]{% 
	  \tt {\tiny [A$\langle$GM]}  
	 ME:  Is the travel agent aware of the dates of your trip? [..] 
	  } 
	   } 
	   } 
	 & \\

    \theutterance \stepcounter{utterance}  

    & \multicolumn{4}{p{0.6\linewidth}}{\cellcolor[rgb]{0.8,1,0.9}{%
	\makecell[{{p{\linewidth}}}]{% 
	  \tt {\tiny [A$\rangle$GM]}  
	 ASIDE: Yes 
	  } 
	   } 
	   } 
	 & & \\

    \theutterance \stepcounter{utterance}  

    & & & \multicolumn{2}{p{0.3\linewidth}}{\cellcolor[rgb]{0.95,0.95,0.95}{%
	\makecell[{{p{\linewidth}}}]{% 
	  \tt {\tiny [GM$|$GM]}  
	 yes 
	  } 
	   } 
	   } 
	 & & \\

    \theutterance \stepcounter{utterance}  

    & & & \multicolumn{2}{p{0.3\linewidth}}{\cellcolor[rgb]{0.95,0.95,0.95}{%
	\makecell[{{p{\linewidth}}}]{% 
	  \tt {\tiny [GM$|$GM]}  
	 Answer for WHEN valid after 1 tries. 
	  } 
	   } 
	   } 
	 & & \\

    \theutterance \stepcounter{utterance}  

    & & & \multicolumn{2}{p{0.3\linewidth}}{\cellcolor[rgb]{0.95,0.95,0.95}{%
	\makecell[{{p{\linewidth}}}]{% 
	  \tt {\tiny [GM$|$GM]}  
	 Answer is correct. 
	  } 
	   } 
	   } 
	 & & \\

    \theutterance \stepcounter{utterance}  

    & & & \multicolumn{2}{p{0.3\linewidth}}{\cellcolor[rgb]{0.95,0.95,0.95}{%
	\makecell[{{p{\linewidth}}}]{% 
	  \tt {\tiny [GM$|$GM]}  
	 End probing 
	  } 
	   } 
	   } 
	 & & \\

\end{supertabular}
}
\caption{An example of the final turn's probing interaction of a Private/Shared episode played by Llama-3.3-70B-Instruct with a programmatic player. The two players' messages are in green and orange, while the grey-colored text comes from the environment itself, represented by the Game Master.}
\label{privateshared_probing_transcript}
\end{figure}

\end{document}